\def\year{2019}\relax
\documentclass[letterpaper]{article} 
\usepackage{aaai19}  
\usepackage{times}  
\usepackage{helvet}  
\usepackage{courier}  
\usepackage{url}  
\usepackage{graphicx}  

\usepackage{epstopdf}
\usepackage{amsmath}
\usepackage{amsfonts}
\usepackage{subcaption}
\usepackage{breqn}
\usepackage{mathtools}

\usepackage{color}

\begin{document}
%
\title{Bias-Variance Trade-Off in Hierarchical Probabilistic Models \\Using Higher-Order Feature Interactions  }
\author{Simon Luo\thanks{This work was accomplished when the first author was at National Institute of Informatics. The code for our implementation of the HBM is available at https://github.com/sjmluo/HBM} \\
        The University of Sydney \\
        Data61, CSIRO \\
        sluo4225@uni.sydney.edu.au
        \And 
        Mahito Sugiyama \\
        National Institute of Informatics \\
        JST, PRESTO \\
        mahito@nii.ac.jp
        }
\maketitle
\begin{abstract}
	Hierarchical probabilistic models are able to use a large number of parameters to create a model with a high representation power. However, it is well known that increasing the number of parameters also increases the complexity of the model which leads to a bias-variance trade-off. Although it is a classical problem, the bias-variance trade-off between \textit{hidden layers} and \textit{higher-order interactions} have not been well studied. In our study, we propose an efficient inference algorithm for the log-linear formulation of the higher-order Boltzmann machine using a combination of Gibbs sampling and annealed importance sampling. We then perform a bias-variance decomposition to study the differences in \textit{hidden layers} and \textit{higher-order interactions}. Our results have shown that using \textit{hidden layers} and \textit{higher-order interactions} have a comparable error with a similar order of magnitude and using \textit{higher-order interactions} produce less variance for smaller sample size.    
\end{abstract}

\section{Introduction}
	Hierarchical machine learning models can be used to identify higher-order feature interactions. They include a wide range of models used in machine learning such as graphical models and deep learning because they can be easily generalized for many different applications. Hierarchical models are widely used as they can use a large number of parameters to create a high representative power for modeling interactions between features. However, tuning towards the optimal model includes a classical machine learning problem known as the \emph{bias-variance trade-off}~\cite{friedman2001elements}. Despite the prevalence of hierarchical models, the bias-variance trade-off for higher-order feature interactions have not been well studied.

\begin{figure}[t]
		\centering
		\begin{subfigure}[b]{0.49\columnwidth}
			\centering
			\includegraphics[width=\textwidth]{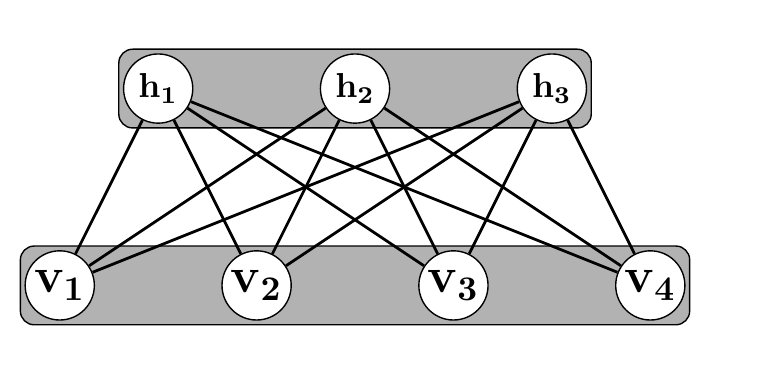}
			\caption{} \label{fig:RBM}
		\end{subfigure}
		\begin{subfigure}[b]{0.49\columnwidth}
			\centering
			\includegraphics[width=\textwidth]{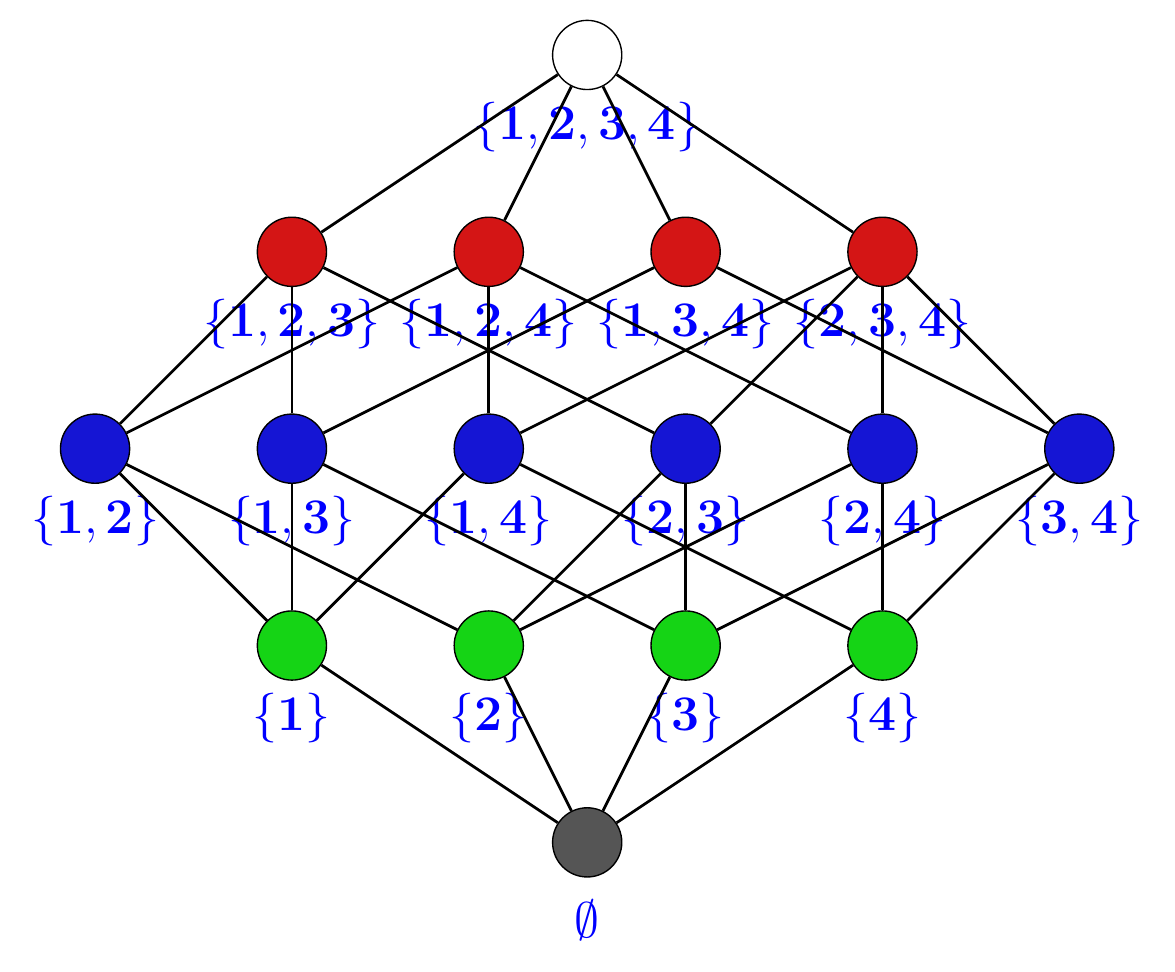}
			\caption{} \label{fig:HBM}
		\end{subfigure}
		\caption{Example of Boltzmann machine modeling high-feature interactions. (\subref{fig:RBM}) A restricted Boltzmann machine with a configuration of 4 visible nodes and 3 hidden nodes. (\subref{fig:HBM}) The outcome space of the Higher-Order Boltzmann machine with 4 visible nodes. The green, blue, red and white nodes show first, second, third and fourth order interactions respectively. The bottom node (black) is used to normalize the Boltzmann machine.}
	\end{figure}

	In this paper, we study the differences in using \textit{hidden layers} and \textit{higher-order interactions} to achieve higher representation power in hierarchical models. In our study, we focus on the Boltzmann Machine (BM)~\cite{ackley1987learning}, one of the fundamental machine learning models. The family of BMs has been used in a wide range of machine learning models including graphical models and deep learning. The Restricted Boltzmann Machine (RBM)~\cite{hinton2012practical} (Figure~\ref{fig:RBM}) and the Higher-Order Boltzmann Machine (HBM)~\cite{sejnowski1986higher,min2014interpretable} (Figure~\ref{fig:HBM}) are fundamental models for learning higher-order feature interactions. However, the two models have different methods for achieving a higher representation power. The RBM is represented by a bipartite graph consisting of two groups, the ``visible layer'' and the ``hidden layer''. The visible layer represents the direct observations of the data, while the hidden layer is used to identify latent features. The bias and the edge weights between the visible nodes and hidden nodes are tuned to model the interaction between the features. The HBM is represented using a finite partially ordered set (poset) using a Hasse diagram~\cite{davey2002introduction,gierz2003continuous}. The poset is used to represent the outcome space of the HBM. A weight is placed on each of the nodes in the outcome space to model the higher-order feature interactions. From a theoretical perspective, both the RBM and HBM are capable models in modeling higher-order feature interactions~\cite{le2008representational}.

	In our study, we empirically perform a \textit{bias-variance decomposition} for both the RBM and HBM to study the total errors of each model from the trade-off between bias and variance. For our study, we use Contrastive Divergence (CD) as the inferencing technique for the RBM. For the HBM, we use the recent information geometric formulation of the HBM by Sugiyama \textit{et al.}~\cite{sugiyama2016information,sugiyama2017tensor} and use gradient descent to maximize the likelihood. The generalized Pythagorean theorem from information geometry enables us to decompose the total error represented as the Kullback--Leibler (KL) divergence into bias and variance terms. Our analysis uses a synthetic dataset with varying features, samples and model complexity. Our contribution includes: 1) A proposal to use a combination of Gibbs sampling and Annealed Importance Sampling (AIS) in inference to overcome the computational and numerical problems in training the HBM. 2) A study which compares the bias-variance trade-off in \textit{hidden layers} and \textit{higher-order interactions}.

	Our results have shown that using \textit{hidden layers} and \textit{higher-order interactions} produce a similar error from the bias and \textit{higher-order interactions} produce less variance for a smaller sample size. For larger datasets, the error from the bias is more dominant, therefore for sufficiently large datasets, \textit{hidden layers} and \textit{higher-order interactions} have a comparable error with a similar order of magnitude.

\section{Formulation} \label{sec:formulation}
	This section presents the hierarchical probabilistic models to analyze the error in modeling higher-order feature interactions. We first introduce the generic Boltzmann machine along with the Restricted Boltzmann Machine (RBM). We then present the information geometry formulation of the log-linear model of hierarchical probability distribution, which includes the family of Boltzmann machines. Finally, we present the Higher-Order Boltzmann Machine (HBM) and our proposed inferencing algorithm.

	\subsection{Boltzmann Machine}
		A \textit{Boltzmann Machine}~\cite{ackley1987learning} is represented using an undirected graph $G = \left( V, E \right)$ with a vertex set $V = \left\{ v_1, v_2, \ldots, v_n, h_1, h_2, \ldots, h_m \right\}$ and an edge set $E \subseteq \left\{ \left\{ x_i, x_j \right\} \mid x_i, x_j \in V \right\}$. Each vertex can be either ``visible'' or ``hidden'', where the visible node represents a direct observation from the dataset and the hidden node represents latent features detected in the model. The state of each vertex is represented by $\mathbf{x} = \left( x_1, x_2, \ldots, x_{n+m} \right) \in \left\{ 0, 1 \right\}^{n+m}$, which is a concatenation of $\mathbf{v} \in \{ 0, 1 \}^n$ and $\mathbf{h} \in \{ 0, 1 \}^m$. The generalized expression for the energy of the joint configuration $\left( \mathbf{v}, \mathbf{h} \right)$ of the network is defined by:
		\begin{align}
			\Phi \left( \mathbf{x}; \mathbf{b}, \mathbf{w} \right) &= - \sum_{i = 1}^{n+m}  x_i b_i - \sum_{i,j = 1}^{n+m} x_i x_j w_{i,j} , \label{eqn:partition_function}
		\end{align}
		where the parameters $\mathbf{b} = \left( b_1, b_2, \ldots, b_{n+m} \right)$ are the biases and $\mathbf{w} = \left( w_{1,2}, w_{1,3}, \ldots, w_{n+m-1, n+m} \right)$ are the weights which are placed on the vertices and the edges respectively. If $\left\{i, j\right\} \notin E$ then $w_{i,j} = 0$ can be used to represent no edge connection between the two vertices. The probability of the current configuration of the entire Boltzmann Machine network $G$ is given by:
		\begin{align}
			p \left( \mathbf{x}; \mathbf{b}, \mathbf{w} \right) &= \frac{\exp \left( - \Phi \left( \mathbf{x}; \mathbf{b}, \mathbf{w} \right) \right)}{Z} , \label{eqn:boltzmann_machine}
		\end{align}
		where $Z$ is the partition function given by:
		\begin{align}
			Z &=\ \ \sum_{\mathclap{\mathbf{x} \in \left\{ 0, 1 \right\}^{n+m}}}\  \exp \left( - \Phi \left( \mathbf{x}; \mathbf{b}, \mathbf{w} \right) \right), \label{eqn:normalization_partition_function}
		\end{align}
		which ensures $\sum_{\mathbf{x} \in \left\{ 0, 1 \right\}^{n+m}} p \left( \mathbf{x}; \mathbf{b}, \mathbf{w} \right) = 1$.

	\subsection{Restricted Boltzmann Machine}
		The Restricted Boltzmann Machine (RBM) is a BM in which the vertices form a bipartite graph. The two groups of nodes are known as the ``visible layer'' and the ``hidden layer''. The visible layer represents the direct observations on the dataset. An activation function is placed on the hidden nodes to model the interactions between features by the following equation:
		\begin{align*}
			p \left( h_j = 1 \middle| \mathbf{v} \right) &= \sigma \left( b_j + \sum_i v_i w_{ij} \right) ,
		\end{align*}
		where $\sigma$ is the sigmoid function.
		We apply the efficient contrastive divergence technique to adjust the weights and biases to maximize the product of probabilities of generating a given dataset by the BM~\cite{Hinton02,Tieleman08}. The updates for the weights and the biases is given by:
		\begin{align*}
			\Delta \mathbf{W} = \epsilon \left( \mathbf{v} \mathbf{h}^T - \mathbf{v}' \mathbf{h}'{}^T \right), \qquad \Delta \mathbf{b} = \epsilon \left( \mathbf{x} - \mathbf{x}' \right).
		\end{align*}
		Where $\mathbf{v}'$, $\mathbf{h}'$, $\mathbf{x}'$ represents the samples reconstructed from the model and $\epsilon$ represents the learning rate.

		Additional hidden layers can be added to create the Deep Boltzmann Machine (DBM) to increase the representation power~\cite{Salakhutdinov09,Salakhutdinov12}. However, this is not analyzed in our study due to the exponential increase in complexity to compute the partition function in Equation~\eqref{eqn:boltzmann_machine}.

	\subsection{Information Geometric Formulation of the Log-Linear Model} \label{sec:loglinearmodel}
		The information geometric log-linear probabilistic model has been introduced by Amari \textit{et al.}~\cite{amari2001information,nakahara2002information,nakahara2006comparison}. Further advances in the log-linear formulation by Sugiyama \textit{et al.}~\cite{sugiyama2016information} has enabled to analytically compute the Fisher information of parameters in hierarchical models. This formulation of the hierarchical model uses a partial order structure to represent the possible model outcomes.

		Here we introduce the log-linear formulation introduced in Sugiyama \textit{et al.}~\cite{sugiyama2016information}. Let $\left( S, \le \right)$ be a \textit{partially ordered set} (\textit{poset})~\cite{gierz2003continuous}, where a \textit{partial order} $\le$ is the relation between elements in a set $S$. The \textit{poset} must satisfy the following three properties for $x, y, z \in S$: (1) $x \le x$ (reflexivity), (2) $x \le y, y \le x \Rightarrow x = y$ (anti-symmetry), and (3) $x \le y, y \le z \Rightarrow x \le z$ (transitivity). We assume that the set $S$ is finite, where $\bot \in S$ and $\bot \le x$, $\forall x \in S$. To be concise, we use $S^{+}$ to denote $S\setminus \left\{ \bot \right\}$.

		The \textit{zeta function} $\zeta : S \times S \rightarrow \left\{ 0, 1 \right\}$ and the \textit{M\"{o}bius function} $\mu : S \times S \rightarrow \mathbb{Z}$ are two functions used to construct the partial order structure. The \textit{zeta function} is defined as:
		\begin{align*}
			\zeta \left( s, x \right) &=
			\begin{cases}
				\ \ 1 \qquad &\text{if } s \le x, \\
				\ \ 0 \qquad &\text{otherwise}.
			\end{cases}
		\end{align*}
		The \textit{M\"{o}bius function} $\mu$ is defined to be the convolution inverse of the \textit{zeta function}, i.e:
		\begin{align*}
			\mu \left( x, y \right) =
			\begin{cases}
				\ \ 1 \qquad &\text{if } x = y, \\
				\ \ -\sum_{x \le s < y} \mu \left( x, s \right) \qquad &\text{if } x < y, \\
				\ \ 0 \qquad &\text{otherwise}.
			\end{cases}
		\end{align*}
		The \textit{log-linear model} on $S$ provides a mapping of the discrete probability distribution to the structured outcome space $\left( S, \le \right)$. Let the probability distribution $P$ denote a probability distribution that assigns a probability $p \left( x \right)$ for each $x \in S$ while satisfying $\sum_{x \in S} p \left( x \right) = 1$. Each probability $p \left( x \right)$ for $x \in S$ is defined as:

		\begin{equation}
			\begin{split}
				\log p \left( x \right) &= \sum_{s \in S} \zeta \left( s, x \right) \theta \left( s \right) = \sum_{s \le x} \theta \left( s \right), \\
				\theta \left( x \right) &= \sum_{s \in S} \mu \left( s, x \right) \log p \left( s \right). \label{eqn:log_linear_theta}
			\end{split}
		\end{equation}
		\begin{equation}
			\begin{split}
				\eta \left( x \right) &= \sum_{s \in S} \zeta \left( x, s \right) p \left( s \right), \\
				p \left( x \right) &= \sum_{s \in S} \mu \left( x, s \right) \eta \left( s \right). \label{eqn:log_linear_etas}
			\end{split}
		\end{equation}
		Sugiyama \textit{et al.}~\cite{sugiyama2017tensor} has shown that the set of distributions $\mathcal{S} = \left\{ P \mid 0 < p \left( x \right) < 1\ \mathrm{and}\ \sum p \left( x \right) = 1 \right\}$ always become a \textit{dually flat Riemannian manifold}. This makes the two functions $\theta$ and $\eta$ a dual coordinate system on $\mathcal{S}$ which is connected through the Legendre transformation.

	\subsection{Higher-Order Boltzmann Machine}
		Higher order interactions in a Boltzmann machine are capable of modeling higher-order feature interactions. However, they are very rarely used in practice due to the high computational cost for inferencing and learning. The log-linear formulation of the Boltzmann machine provides an elegant representation of the outcome space. This formulation allows any parameters to be included or removed from $S^{+}$. For a given Boltzmann machine $S(B) = 2^{V}$ with $V = \left\{ 1, 2, \ldots, n \right\}$, the energy function of the $k$th order Boltzmann machine is defined as:
		\begin{multline*}
			\Phi \left( \mathbf{x}; \mathbf{b}, \mathbf{w} \right) = -\sum_{{i_1} \in V} b_{{i_1}} x_{{i_1}} - \sum_{{i_1},{i_2} \in V} w_{{i_1}{i_2}} x_{i_1} x_{i_2} \\
			- \sum_{i_1,i_2,i_3 \in V} w_{{i_1}{i_2}{i_3}} x_{i_1} x_{i_2} x_{i_3} \\
			- \dots - \sum_{{i_1},{i_2},\ldots,{i_k} \in V} w_{{i_1}, {i_2}, \ldots, {i_k}} x_{i_1} x_{i_2} \ldots x_{i_k},
		\end{multline*}
		Sugiyama \textit{et al.}~\cite{sugiyama2016information,sugiyama2017tensor} have shown that the log-linear model can be used to represent the family of Boltzmann Machines. A \textit{submanifold} of $\mathcal{S}$ can be used to represent the set of Gibbs distribution of the BM $B$ given by $\mathcal{S} \left( B \right) = \left\{ P \in \mathcal{S} \mid \theta \left( x \right) = 0, \forall x \notin B \right\}$. The Gibbs distribution in Equation~\eqref{eqn:boltzmann_machine} directly corresponds to the log-linear model in Equation~\eqref{eqn:log_linear_theta} by:
		\begin{equation}
			\begin{split}
			\log p \left( x \right) &= \sum_{s \in B} \zeta \left( s, x \right) \theta \left( s \right) - \psi \left( \theta \right), \\
			\psi \left( \theta \right) &= - \theta \left( \bot \right) = \log Z, \label{eqn:log_linear_boltzmann}
			\end{split}
		\end{equation}
		where magnitude of $\theta \left( x \right)$ corresponds to the model parameters which model the order of interactions in the Boltzmann machine $B = \left\{ x \in S^{+} \mid \left| x \right| = 1 \mathrm{\ or\ } x \in E \right\}$, that is; $\theta \left( x \right) = b_i$ if $| x | = 1$ and $\theta \left( x \right) = w_{x_{ij}}$ if $\left| x \right| = 2$. The log-linear formulation of the Boltzmann machine shown in Equation~\eqref{eqn:log_linear_boltzmann} can be extended to be a $k$th order Boltzmann machine by $B = \left\{ x \in S^{+} \mid \left| x \right| \le k \right\}$.

		\subsubsection{Inferencing Algorithm}
			The log-linear formulation of the Boltzmann machine can be trained by minimizing the KL (Kullback-Leibler) divergence to approximate a given empirical distribution $\hat{P}$:
			\begin{align}
				\min_{P_B \in \mathcal{S} \left( B \right)} D_{\mathrm{KL}} \left( \hat{P}, P_B \right) &= \min_{P_B \in \mathcal{S} \left( B \right)}\ \ \ \sum_{\mathclap{P_B \in \mathcal{S} \left( B \right)}}\ \hat{p} \left( s \right) \log \frac{\hat{p} \left( s \right)}{p_B \left( s \right)}. \label{eqn:KL_diveregence_BM}
			\end{align}
			This is equivalent to maximizing the log-likelihood $L \left( P_B \right) = N \sum_{s \in S} \hat{p} \left( s \right) \log p_{B} \left( s \right)$.
			The gradient is obtained as,
			\begin{align*}
				\frac{\partial}{\partial \theta_B \left( x \right)} D_{\mathrm{KL}} \left( \hat{P}, P_B \right) = \frac{\partial}{\partial \theta_B \left( x \right)} \sum_{s \in S} \hat{p} \left( s \right) \log p_B \left( s \right)
			\end{align*}
			\begin{multline*}
				= \frac{\partial}{\partial \theta_B \left( x \right)} \sum_{s \in S} \left( \hat{p} \left( s \right) \sum_{\bot < u \le s} \theta_B \left( u \right) \right) \\
				- \frac{\partial}{\partial \theta_B \left( x \right)} \psi \left( \theta_B \right) \sum_{s \in S} \hat{p} \left( s \right)
			\end{multline*}
			\begin{align*}
				= \hat{\eta} \left( x \right) - \eta_B \left( x \right).
			\end{align*}
			However, $\eta_{B} \left( x \right)$ is computationally expensive to compute because it requires to compute all values of $P_B$. We propose to use a combination of Gibbs sampling and Annealed Important Sampling (AIS)~\cite{neal2001annealed,salakhutdinov2008learning} to approximate the distribution of $P_B$.

			\subsubsection{Gibbs sampling for $\eta_B$}
				Gibbs sampling~\cite{geman1984stochastic} is a Markov Chain Monte Carlo (MCMC) algorithm which approximates a multivariate probability distribution. It fixes all the other model parameters and updates each of the model parameters one-by-one until convergence:
				\begin{align*}
					P \left( \mathbf{x}_i =1 \middle| \mathbf{x}_{-i}; \theta \right) = \frac{P \left( \mathbf{x}; \theta \right)}{P \left( \mathbf{x}_{-i}; \theta \right)} \propto P \left( \mathbf{x}; \theta \right),
				\end{align*}
				where $\mathbf{x}_{-i}= (x_{1}, \ldots, x_{i-1}, x_{i+1}, \ldots, , x_{n})$. We apply Gibbs sampling to generate samples for $\mathbf{x} = \left( x_1, \ldots, x_n \right)$. By generating samples, we are able to approximate the un-normalized distribution $f^{*}$. The un-normalized probability distribution is proportional to the normalized distribution by a constant $Z$, i.e. $P = \frac{1}{Z} f^{*} \propto f^{*}$. We will later provide a solution using AIS to approximate the normalization constant $Z$. To update each $\mathbf{x}_i$, we use the difference between the energy functions in each node.
				\begin{align*}
				    &\Delta \Phi_i \left( \mathbf{x} ; \theta \right) = \Phi_i \left( \mathbf{x}_{x_i = 0}; \theta \right) - \Phi_i \left( \mathbf{x}_{x_i = 1}; \theta \right) \\
					=\ &- C \log \left( P \left( \mathbf{x}_{x_i = 0}; \theta \right) \right) - \left(- C \log \left( P \left( \mathbf{x}_{x_i = 1}; \theta \right) \right) \right) \\
					=\ &C \log \left( P \left( \mathbf{x}_{x_i = 1}; \theta \right) - C \log \left( 1 - P \left( \mathbf{x}_{x_i = 1}; \theta \right) \right) \right),
				\end{align*}
				where $C$ represents the constant in the Boltzmann distribution. Rearranging the equation to solve for $P \left( \mathbf{x}_{x_i = 1}; \theta \right) $, we have
				\begin{align*}
					\exp \left( - \frac{\Delta \Phi_i \left( \mathbf{x}; \theta \right)}{C} \right) &= \frac{1 - P \left( \mathbf{x}_{x_i = 1}; \theta \right)}{P \left( \mathbf{x}_{x_i = 1}; \theta \right)}, \\
					P \left( \mathbf{x}_{x_i = 1}; \theta \right) &= \frac{\exp \left( \Delta \Phi_i \left( \mathbf{x}; \theta \right) / C \right)}{1 + \exp \left( \Delta \Phi_i \left( \mathbf{x}; \theta \right) / C \right)}.
				\end{align*}
				The term with the change in energy is calculated by:
				\begin{align*}
					&\exp{ \left( \frac{\Delta \Phi_i \left( \mathbf{x} ; \theta \right)}{C} \right) } \\
					=\ &\exp{ \left( \frac{1}{C} \left[ \Phi_i \left( \mathbf{x}_{\mathbf{x}_i = 0}; \theta \right) - \Phi_i \left( \mathbf{x}_{\mathbf{x}_i = 1}; \theta \right) \right] \right) } \\
					=\ &\exp{ \biggl( - \log P \left( \mathbf{x}_{\mathbf{x}_i = 0}; \theta \right) - \left(- \log P \left( \mathbf{x}_{\mathbf{x}_i = 1}; \theta \right) \right)  \biggr) } \\
					=\ &\exp{ \left( \sum_{s \in S} \zeta \left( s, \mathbf{x}_{\mathbf{x}_i = 1} \right) \theta \left( s \right) - \sum_{s \in S} \zeta \left( s, \mathbf{x}_{\mathbf{x}_i = 0} \right) \theta \left( s \right) \right) }.
				\end{align*}
				A set of $M$ samples can be generated to approximate $\eta_B$ by using Equation~\eqref{eqn:log_linear_etas} by empirically estimating the distribution of $P^*_B$. The overall run-time for each interaction for the Gibbs sampling step is $\mathcal{O} \left( M \left| \mathcal{S}(B) \right| ^2 \right)$.

			\subsubsection{Annealed Importance Sampling to Approximate $P_B$}
				We propose to use AIS~\cite{neal2001annealed,salakhutdinov2008learning} to overcome the numerical problems in computing the normalization parameter in the partition function $Z$. By inspecting Equation~\eqref{eqn:normalization_partition_function}, we can identify a number of problems in computing $Z$. Firstly, it is clear that the value of $Z$ is extremely large because it takes the sum of the exponential of all energy functions. The large value of $Z$ often creates numerical problems for most implementation. Secondly, computing the energy function $\Phi \left( x \right)$ for all nodes is extremely computationally expensive. AIS provides a solution to approximate the value of $\log \left(Z \right)$ without having to compute $Z$ or evaluate the energy function $\Phi \left( x \right)$.

				AIS approximates the normalization constant by tracking the gradual changes of an MCMC transition $T_k \left( \mathbf{x}_{n+1} \middle| \mathbf{x}_{n} \right)$ operation such as Gibbs sampling. AIS uses a sequence of intermediate probability distributions to evaluate the importance weight $w^{\left( i \right)}_{\mathrm{AIS}}$ which is an estimation of the ratio between the first and last distribution.

				For our study, we use one of the most prevalent methods to generate a sequence of intermediate probability distributions for $k=0, \ldots, K$ by using the following geometric property,
				\begin{align*}
					f_{k} \left( x \right) \propto f^{*}_{0} \left( x \right)^{1 - \beta_k} f^{*}_{k} \left( x \right)^{\beta_k}
				\end{align*}
				where $0 = \beta_0 < \beta_1 < \ldots < \beta_K = 1$. There have been several other more advanced techniques to model the path of the intermediate distributions~\cite{grosse2013annealing}. However, this is not the focus of our study and can be subjected to further study in future work. The AIS weight $w_{\mathrm{AIS}}$ can be calculated by using
				\begin{align*}
					w_{\mathrm{AIS}}^{\left( i \right)} &= \frac{ f^{*}_{1} \left( x_{1} \right) }{ f^{*}_{0} \left( x_{1} \right) } \frac{ f^{*}_{2} \left( x_{2} \right) }{ f^{*}_{1} \left( x_{2} \right) } \ldots \frac{ f^{*}_{K-1} \left( x_{K-1} \right) }{ f^{*}_{K-2} \left( x_{K-1} \right) } \frac{ f^{*}_{K} \left( x_{K} \right) }{ f^{*}_{K-1} \left( x_{K} \right) },
				\end{align*}
				where $f^{*}$ denotes a function to calculate the un-normalized probability distribution. After completing $M$ runs of AIS, the ratio of the first and final constant of the partition function can be estimated as
				\begin{align}
					\frac{Z_K}{Z_0} \approx \frac{1}{M} \sum_{i=1}^{M} w_{\mathrm{AIS}}^{\left( i \right)} = \hat{r}_{\mathrm{AIS}} \label{eqn:AIS_ratio}
				\end{align}
				\textit{Neal}~\cite{neal2001annealed,neal2005estimating} has theoretically shown that the $\mathrm{Var} \left( \hat{r}_{\mathrm{AIS}} \right) \propto 1 / MK$. For practical implementations Equation~\eqref{eqn:AIS_ratio} should be in log scale to avoid numerical problems. The standard form is shown here for conciseness. From Equation~\eqref{eqn:AIS_ratio}, the final $\log Z $ can be estimated without computing $Z$ if $Z_0$ is known. Then $\log Z_0$ can be calculated efficiently if we initialize $P_B$ uniformly, i.e. for HBM, $ \theta \left( \bot \right) = - \log Z = N \log \left(2 \right) $ and $ \theta \left( x \right) = 0$, $\forall x\in \mathcal{S}^+ $.			

		\begin{figure}[t]
			\includegraphics[width=\columnwidth]{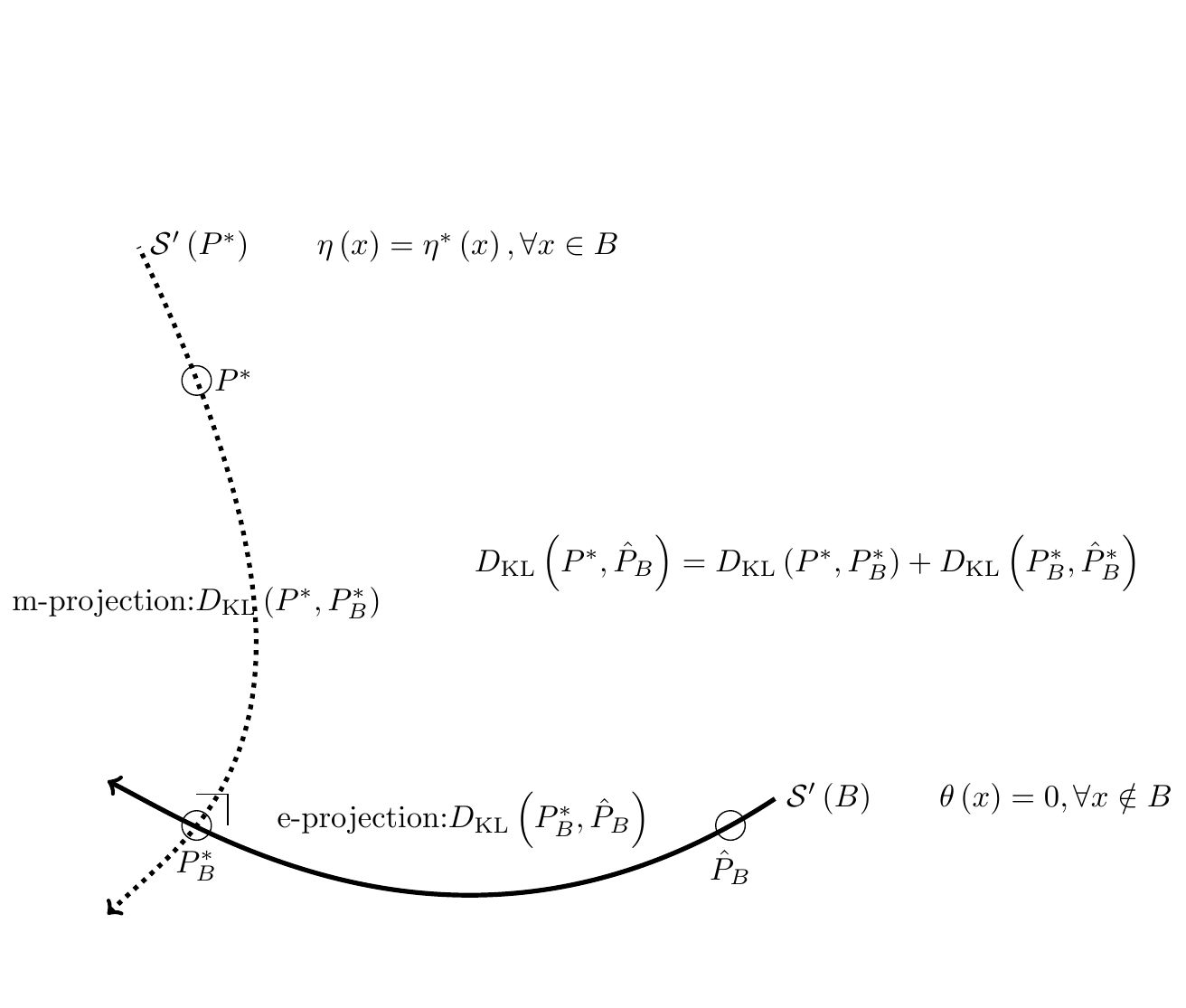}
			\caption{An illustration of the decomposition of the bias and variance.} \label{fig:bias_variance_decomposition}
		\end{figure}

\section{Experiments} \label{sec:experiment}
	Here we present the main results of the paper. This section presents the formulation of the \textit{bias-variance decomposition}, the set-up of the experiment and the experimental results and discussion.

        \begin{figure*}[t]
          \centering
          \begin{subfigure}[b]{0.33\textwidth}
              \centering
              \includegraphics[width=\textwidth]{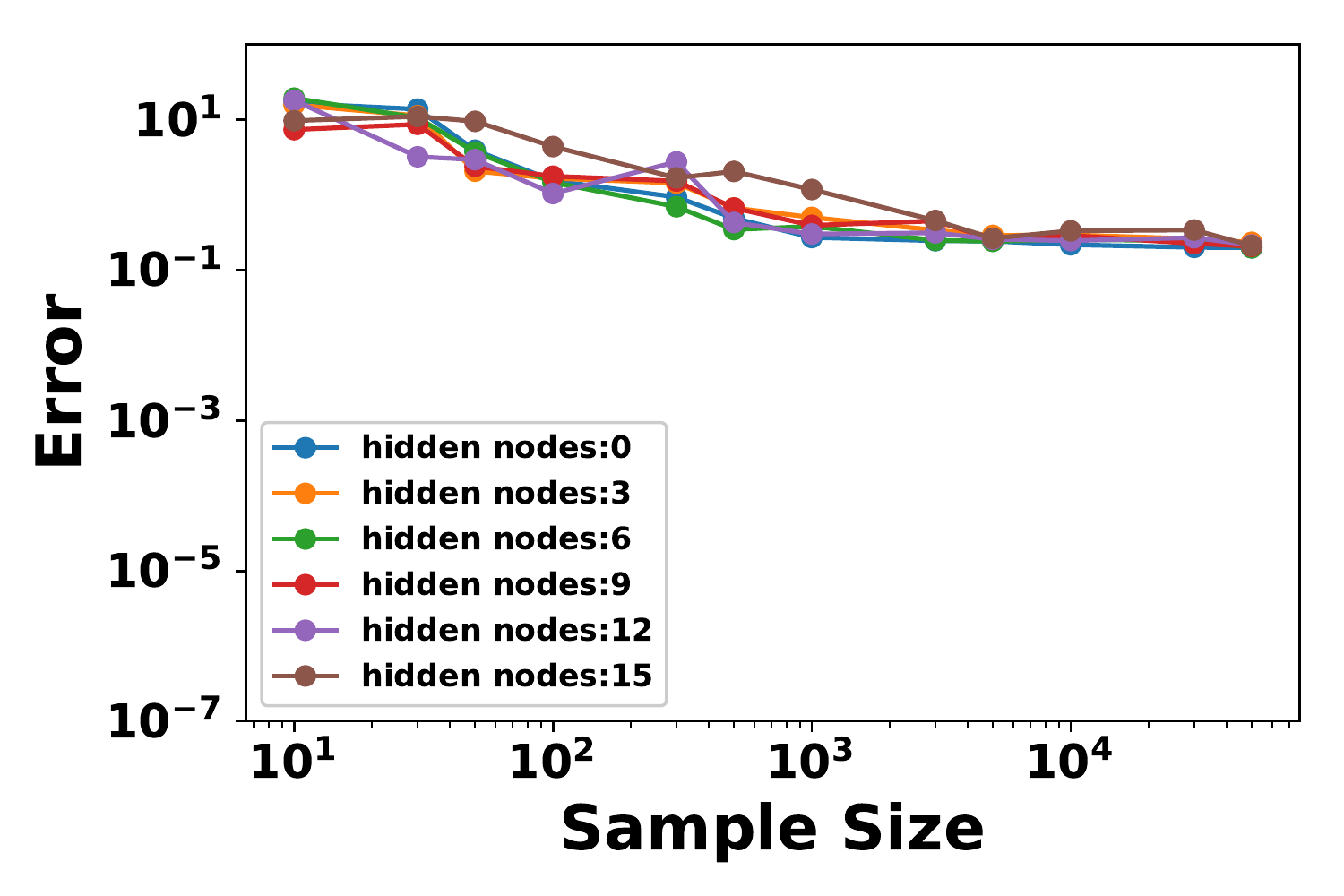}
              \caption{Total Error}
          \end{subfigure}
          \begin{subfigure}[b]{0.33\textwidth}
              \centering
              \includegraphics[width=\textwidth]{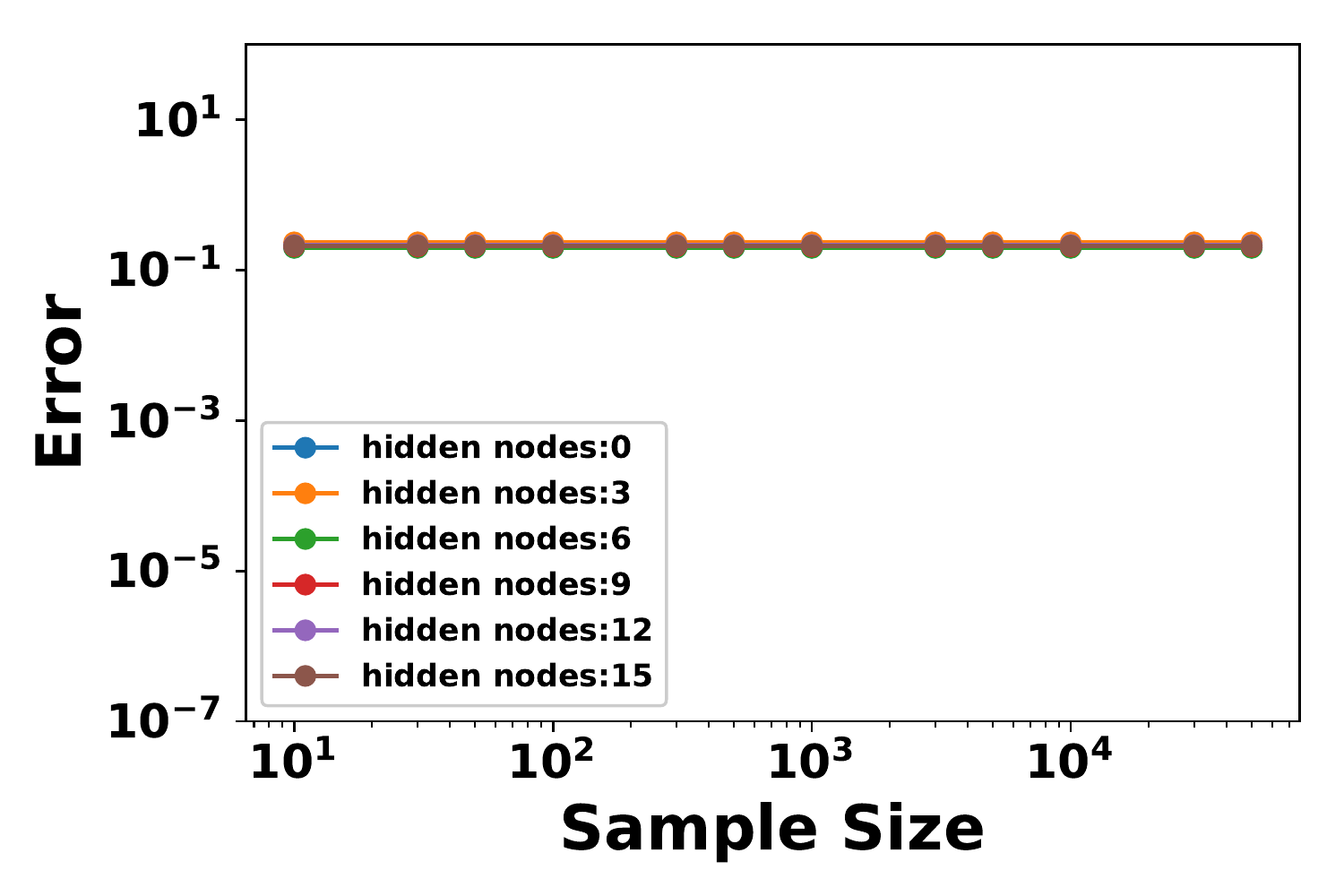}
              \caption{Bias}
          \end{subfigure}
          \begin{subfigure}[b]{0.33\textwidth}
              \centering
              \includegraphics[width=\textwidth]{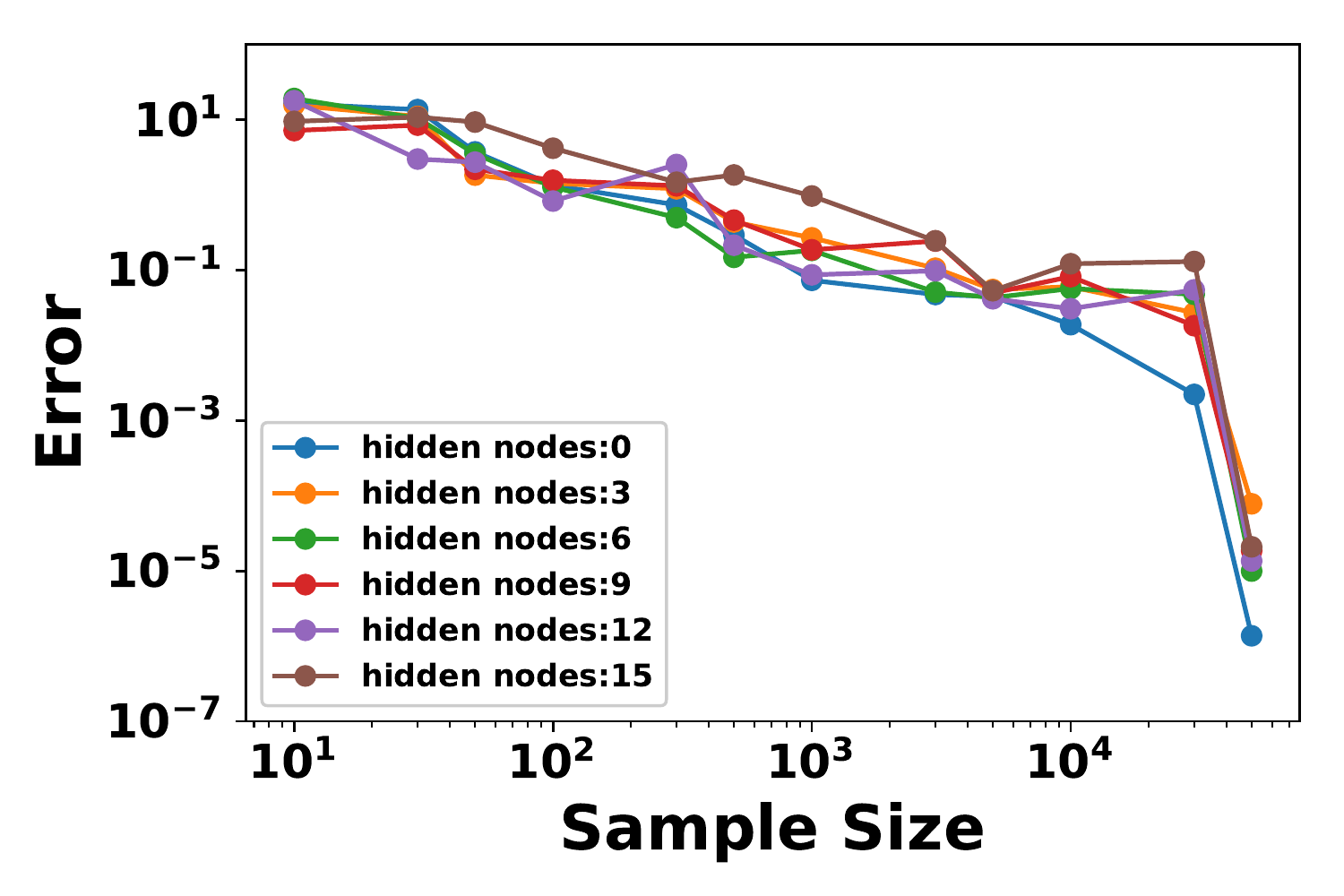}
              \caption{Variance}
          \end{subfigure}
          \caption{Empirical evaluation of the error generated from the bias and variance of the RBM} \label{fig:RBM_error_plot}
        \end{figure*}
        
        \begin{figure*}[t]
          \centering
          \begin{subfigure}[b]{0.33\textwidth}
              \centering
              \includegraphics[width=\textwidth]{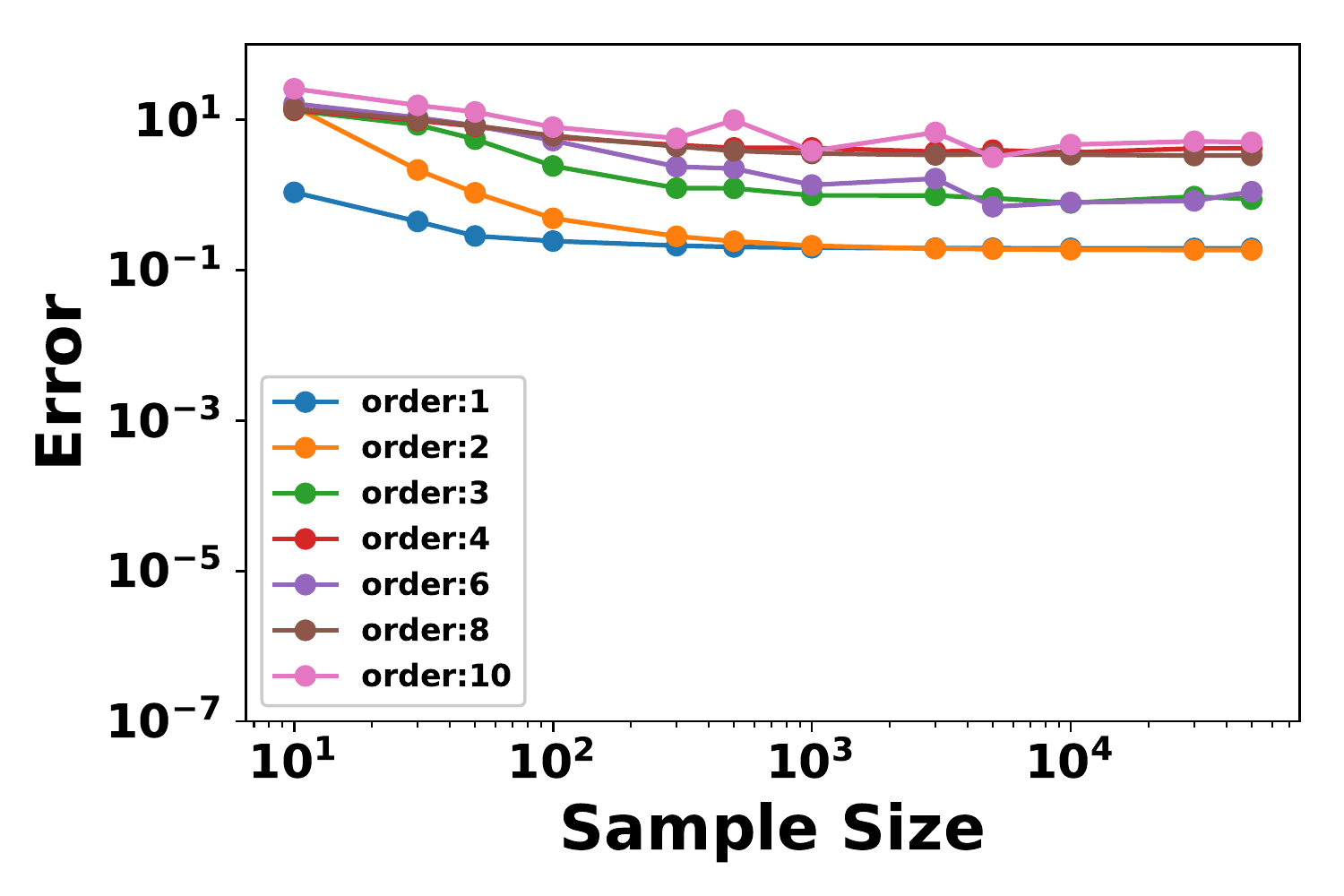}
              \caption{Total Error}
          \end{subfigure}
          \begin{subfigure}[b]{0.33\textwidth}
              \centering
              \includegraphics[width=\textwidth]{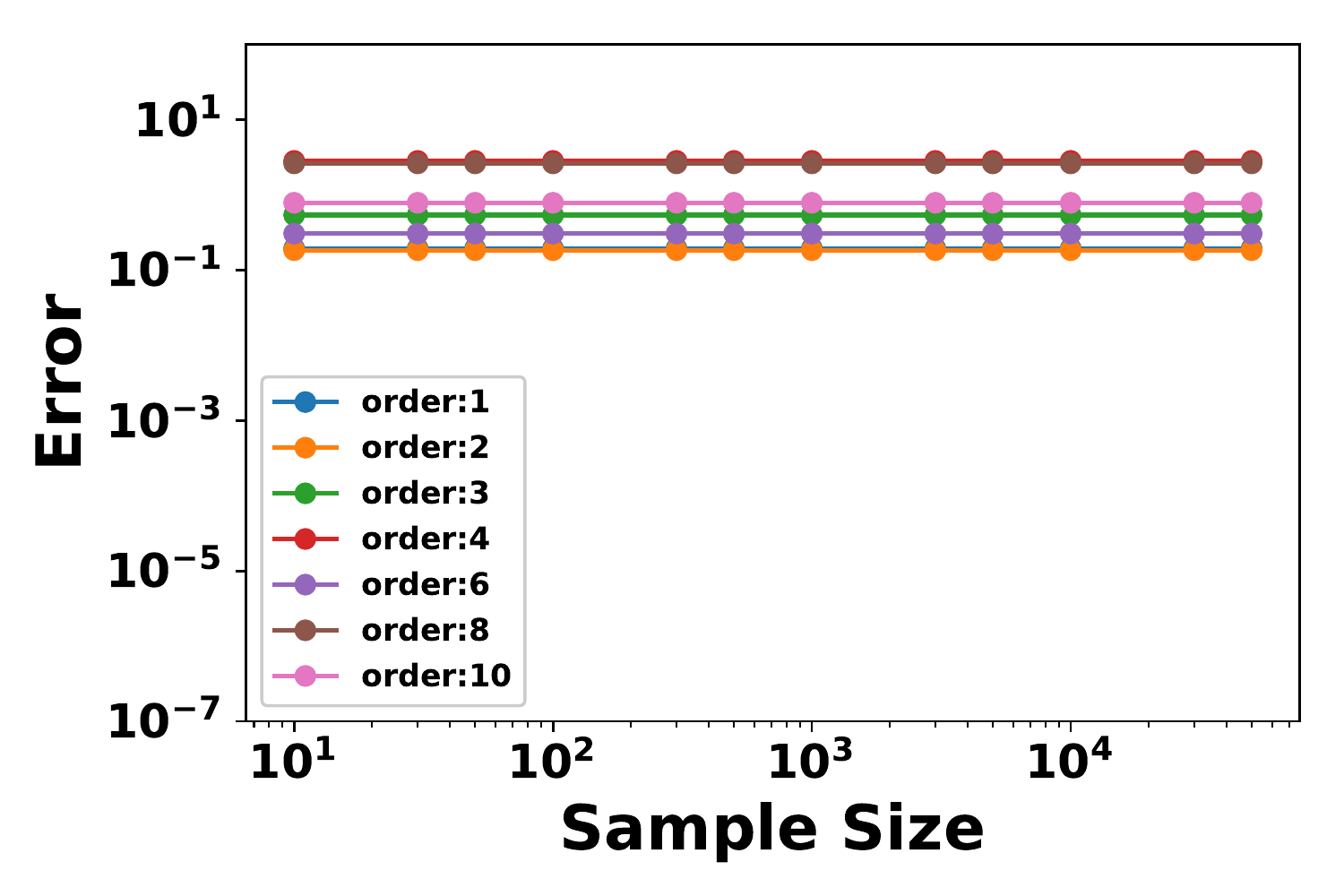}
              \caption{Bias}
          \end{subfigure}
          \begin{subfigure}[b]{0.33\textwidth}
              \centering
              \includegraphics[width=\textwidth]{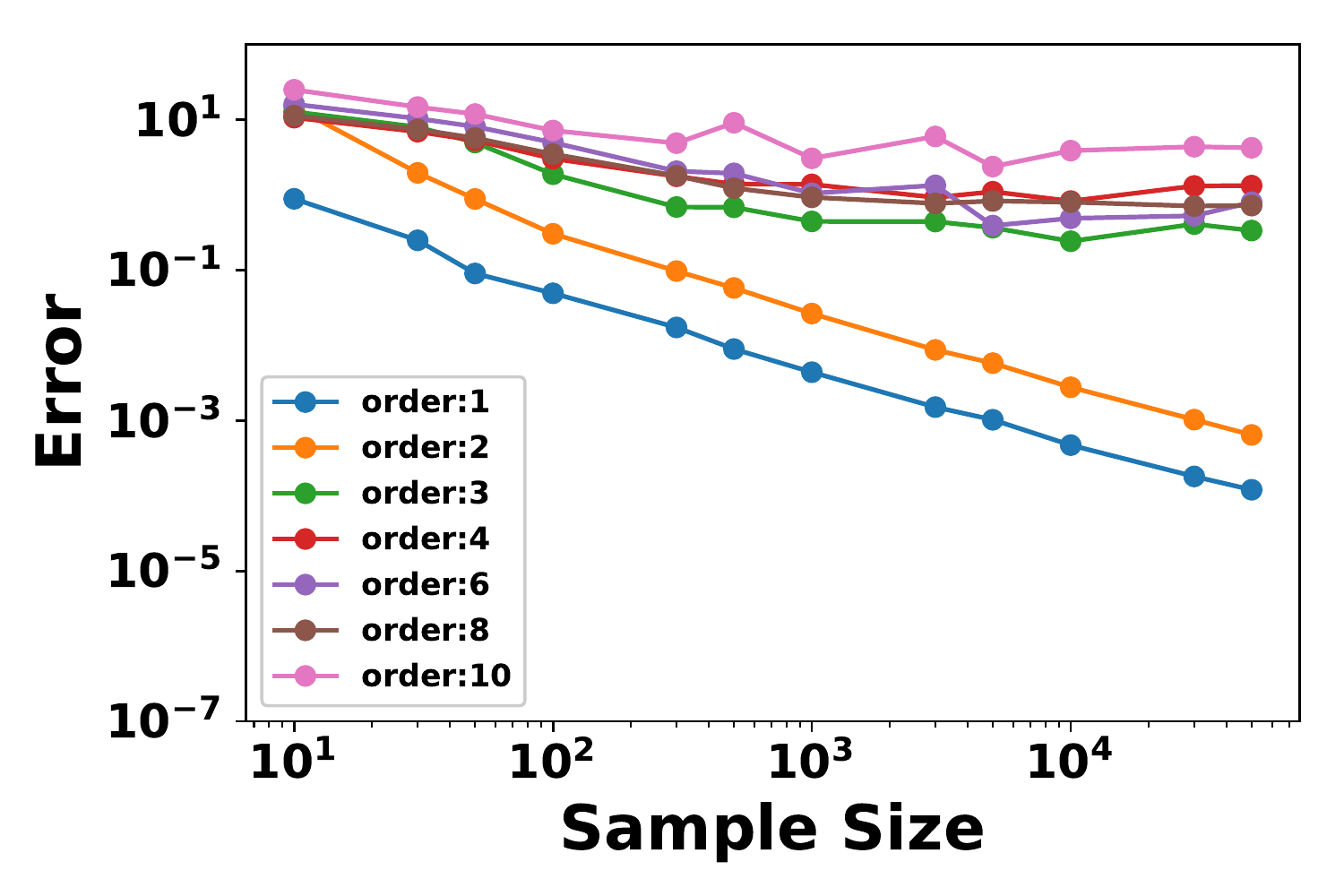}
              \caption{Variance}
          \end{subfigure}
          \caption{Empirical evaluation of the error generated from the bias and variance for the HBM} \label{fig:HBM_error_plot}
        \end{figure*}

	\subsection{Bias-Variance Decomposition}
		We use a \textit{bias-variance decomposition} to compare the behavior of the higher-order feature interactions between the RBM and HBM by varying the complexity of the model. We focus on the expectation of the KL divergence $\mathbf{E} [ D_{\mathrm{KL}} ( P^{*}, \hat{P_B} )]$ from the true (unknown) distribution $P^*$ to the maximum likelihood estimation (MLE) $\hat{P}_B$ of the empirical distribution $\hat{P}$ by a Boltzmann machine with the parameter set $B$. This term represents the total error in the model accumulated from the bias and variance in the model.

		The KL divergence for probabilities in the exponential family can be decomposed into bias and variance using its information geometric properties. We can calculate the true variance of the model by replacing $\hat{P}$ with $P^*$ in Equation~\eqref{eqn:KL_diveregence_BM}. The bias and variance can be separated into two components which are orthogonal as illustrated in Figure~\ref{fig:bias_variance_decomposition}. Using this decomposition of the bias and variance, the total error in the model can be calculated using the \textit{Generalized Pythagorean Theorem},
		\begin{align*}
				&\mathbf{E} \left[ D_{\mathrm{KL}}  \left( P^{*}, \hat{P}_{B} \right) \right] \\
                =\ &\mathbf{E} \left[ D_{\mathrm{KL}} \left( P^{*}, P^{*}_{B} \right) \right] + \mathbf{E} \left[ D_{\mathrm{KL}} \left( P^{*}_{B}, \hat{P}_{B} \right) \right]\\
				=\ &D_{\mathrm{KL}} \left( P^{*}, P^{*}_{B} \right) + \mathbf{E} \left[ D_{\mathrm{KL}} \left( P^{*}_{B}, \hat{P}_{B} \right) \right] \\
				=\ &\underbrace{D_{\mathrm{KL}} \left( P^{*}, P^{*}_{B} \right)}_{\mathrm{bias}} + \underbrace{\mathrm{var} \left( P^{*}_B, B \right)}_{\mathrm{variance}}.
		\end{align*}

        \begin{figure*}[p]
          \centering
          \begin{subfigure}[b]{0.245\textwidth}
              \centering
              \includegraphics[width=\textwidth]{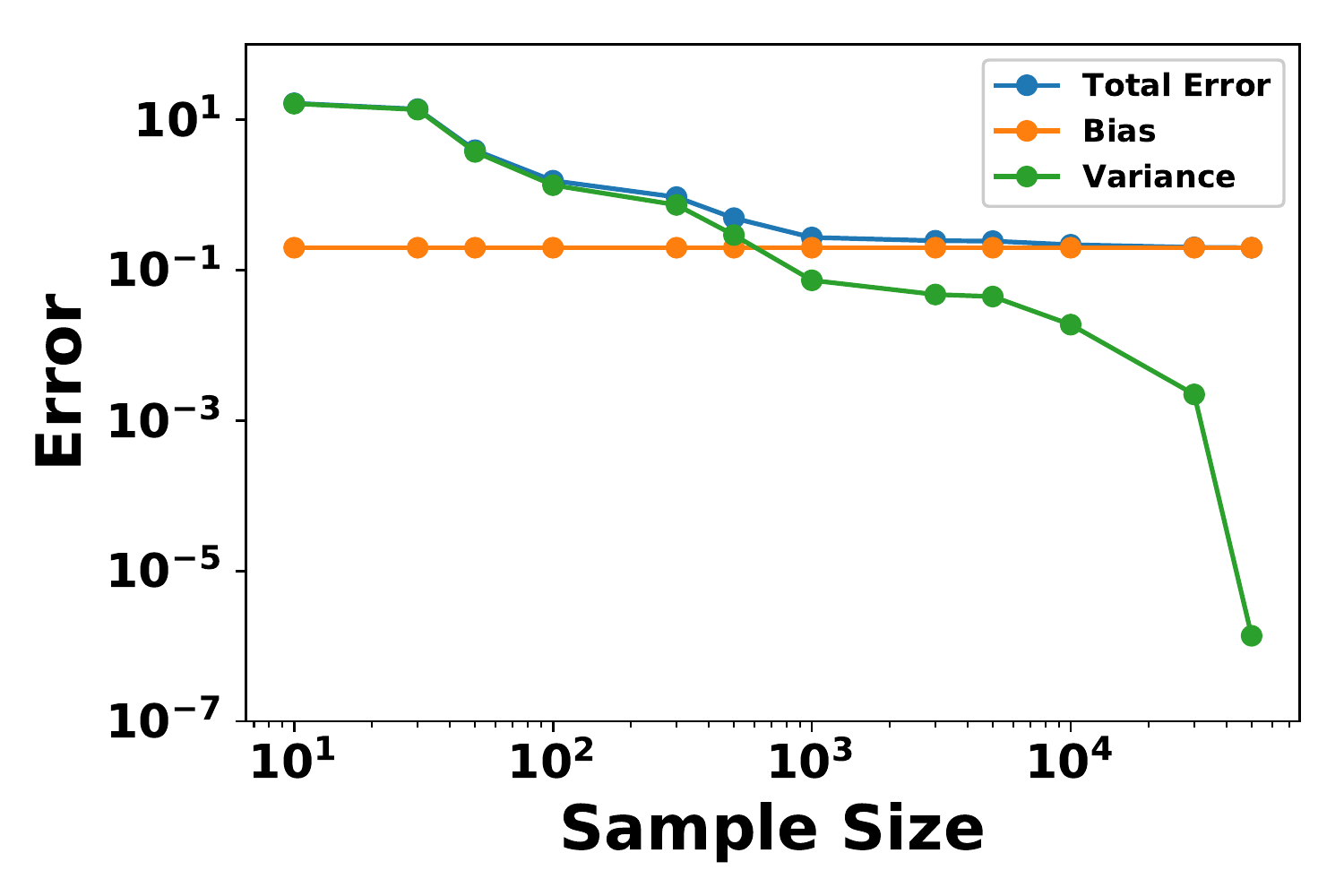}
              \caption{0 Hidden Nodes}
          \end{subfigure}
          \begin{subfigure}[b]{0.245\textwidth}
              \centering
              \includegraphics[width=\textwidth]{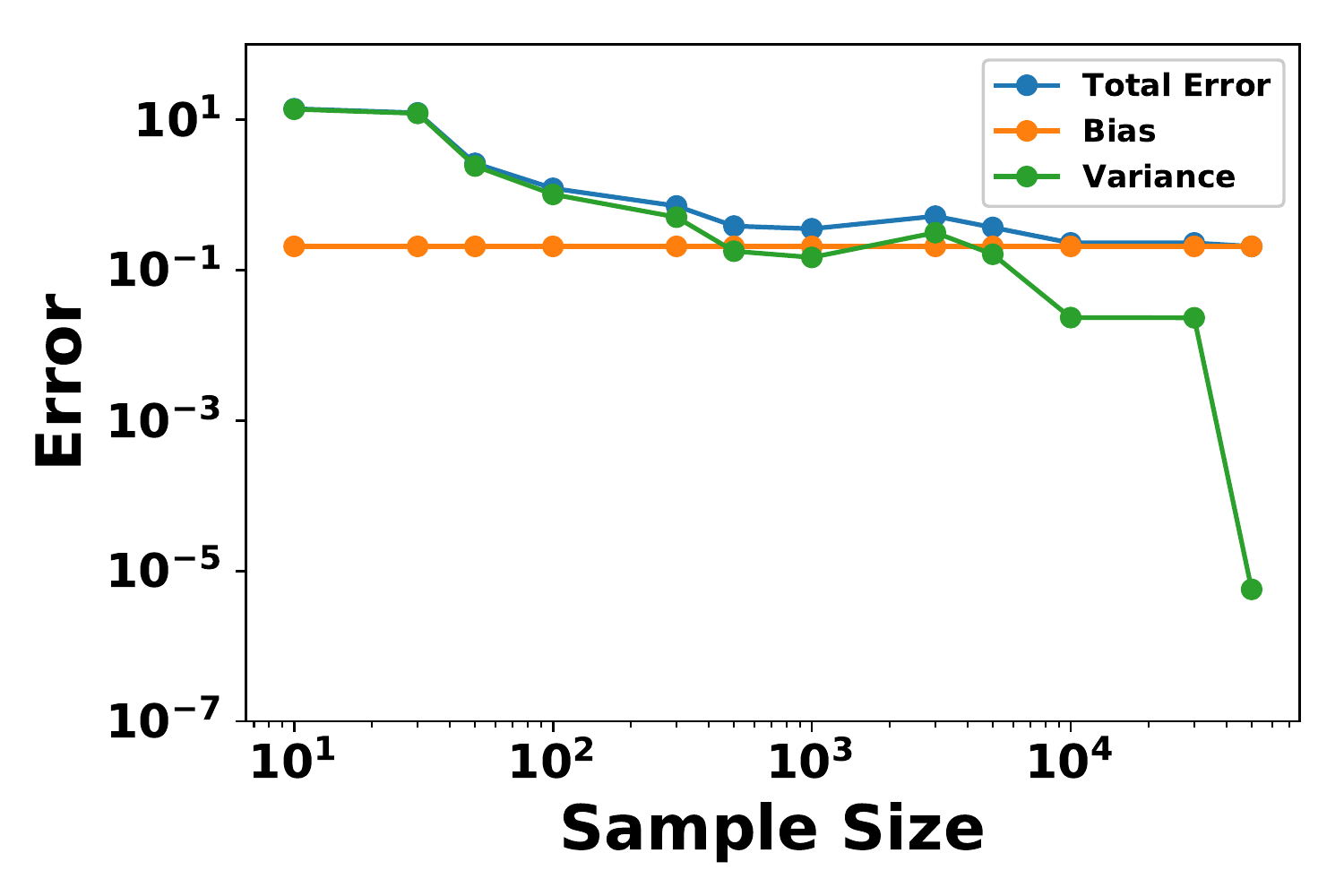}
              \caption{5 Hidden Nodes}
          \end{subfigure}
          \begin{subfigure}[b]{0.245\textwidth}
              \centering
              \includegraphics[width=\textwidth]{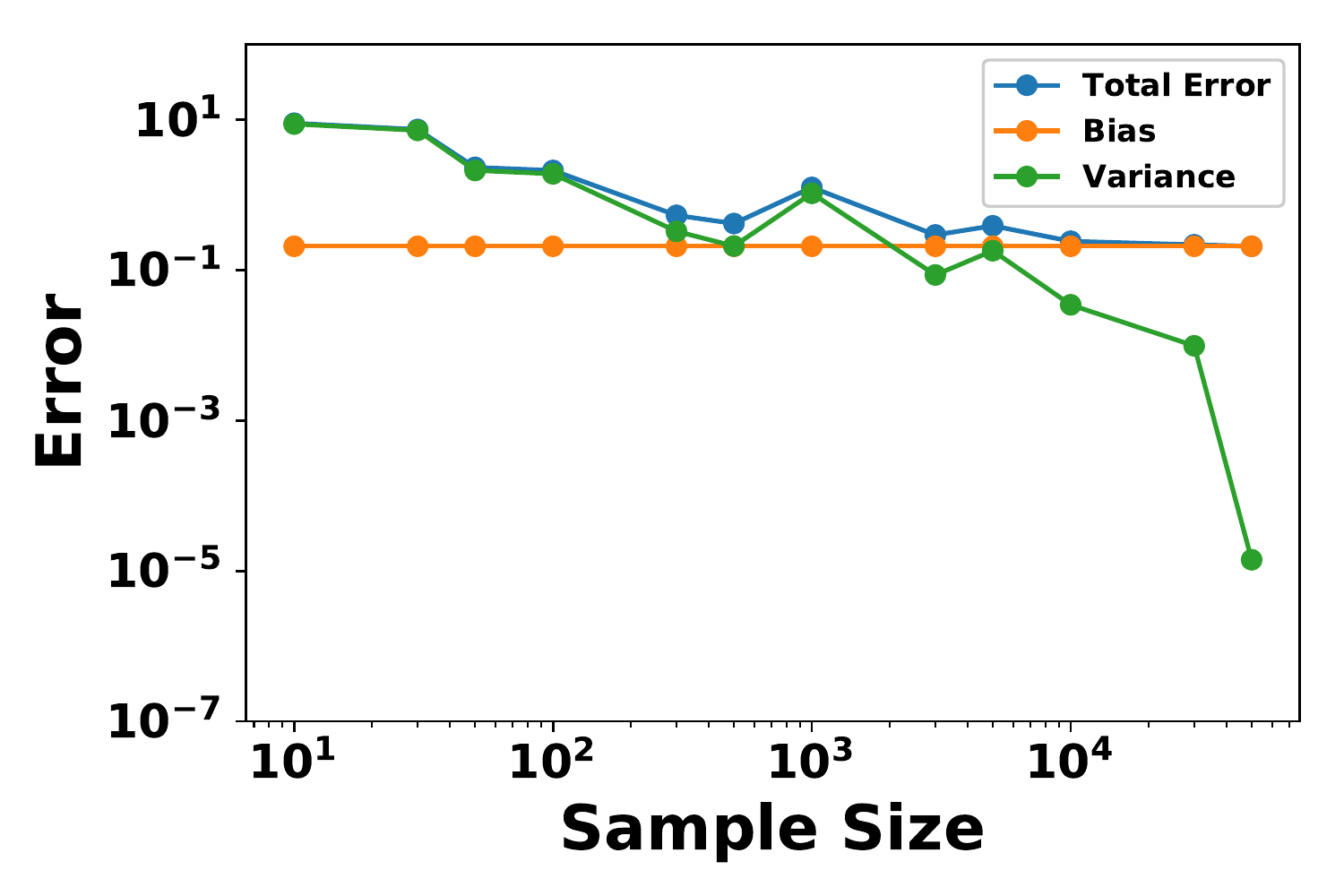}
              \caption{10 Hidden Nodes}
          \end{subfigure}
          \begin{subfigure}[b]{0.245\textwidth}
              \centering
              \includegraphics[width=\textwidth]{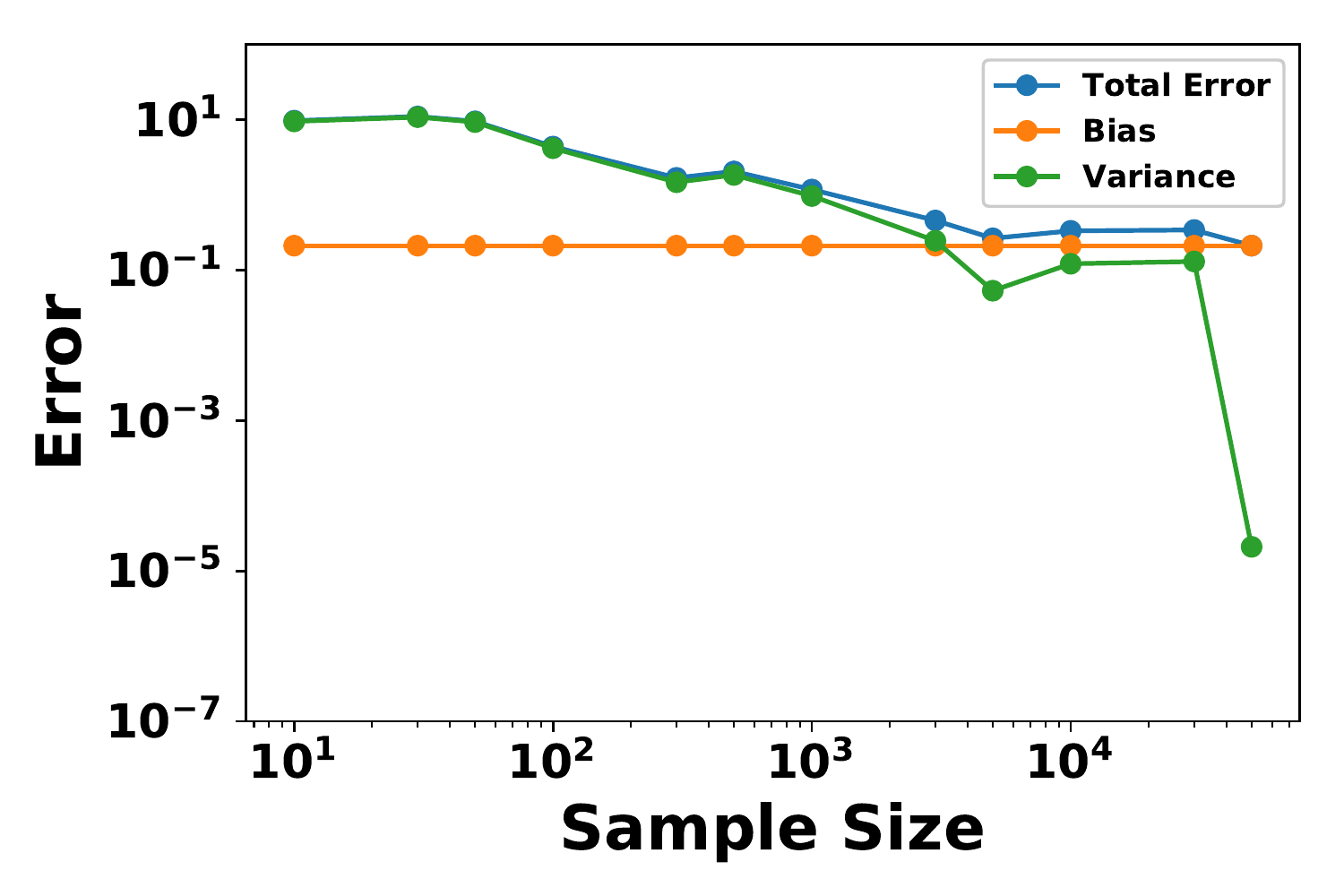}
              \caption{15 Hidden Nodes}
          \end{subfigure}
          \caption{Empirical evaluation of the error generated from the bias and variance for varying hidden nodes in the RBM} \label{fig:RBM_sample_size}
        \end{figure*}

        \begin{figure*}[p]
          \centering
          \begin{subfigure}[b]{0.245\textwidth}
              \centering
              \includegraphics[width=\textwidth]{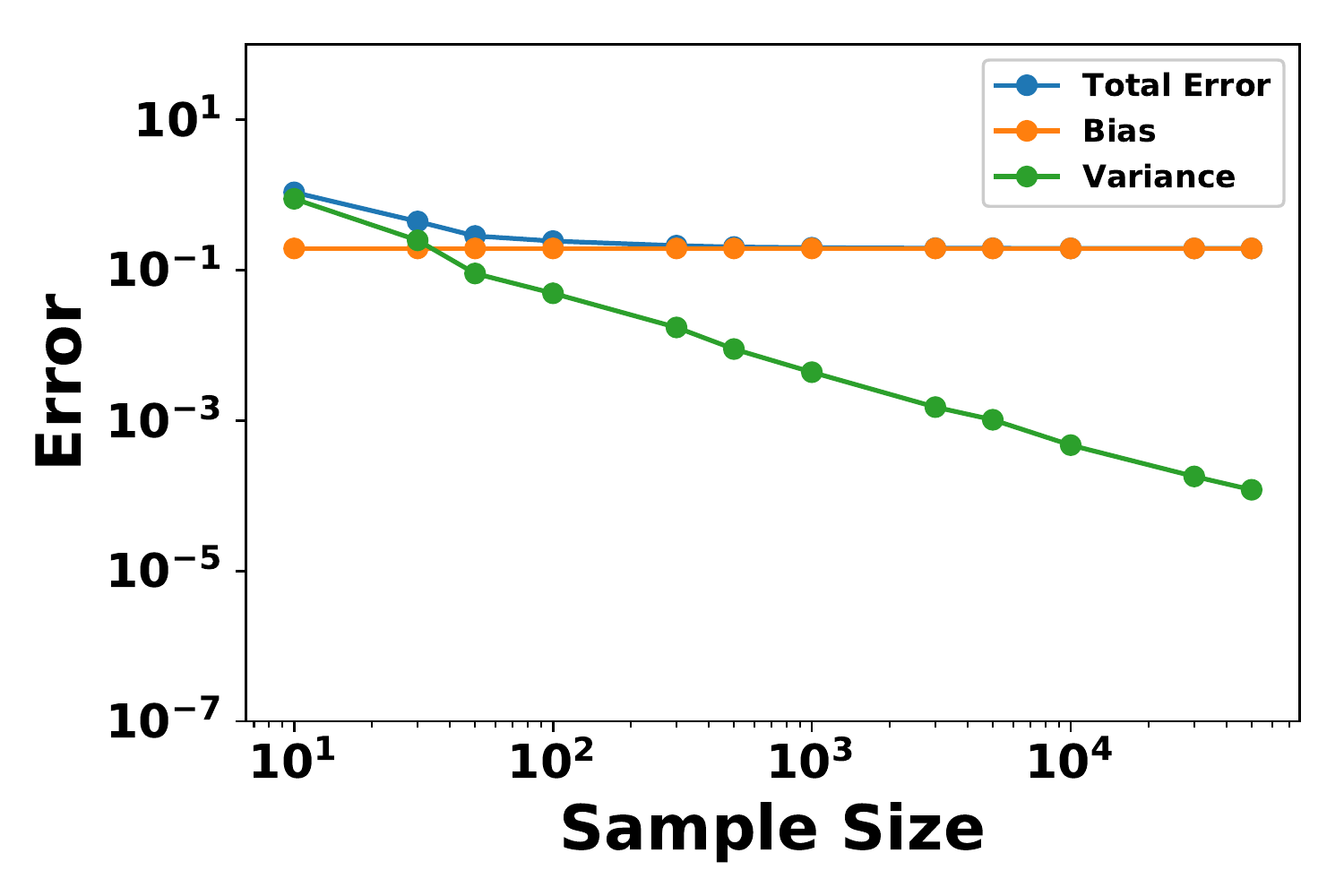}
              \caption{Order 1}
          \end{subfigure}
          \begin{subfigure}[b]{0.245\textwidth}
              \centering
              \includegraphics[width=\textwidth]{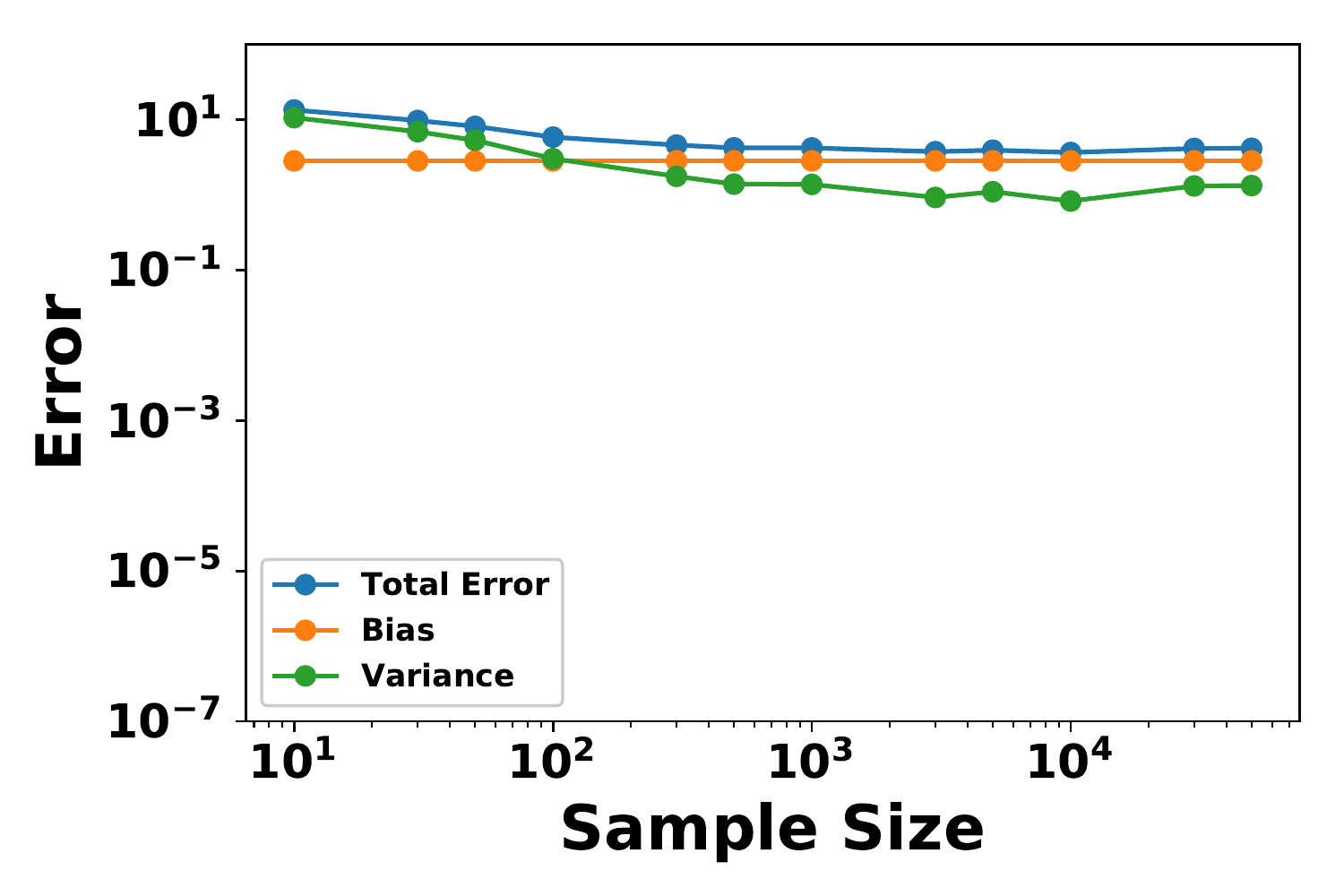}
              \caption{Order 4}
          \end{subfigure}
          \begin{subfigure}[b]{0.245\textwidth}
              \centering
              \includegraphics[width=\textwidth]{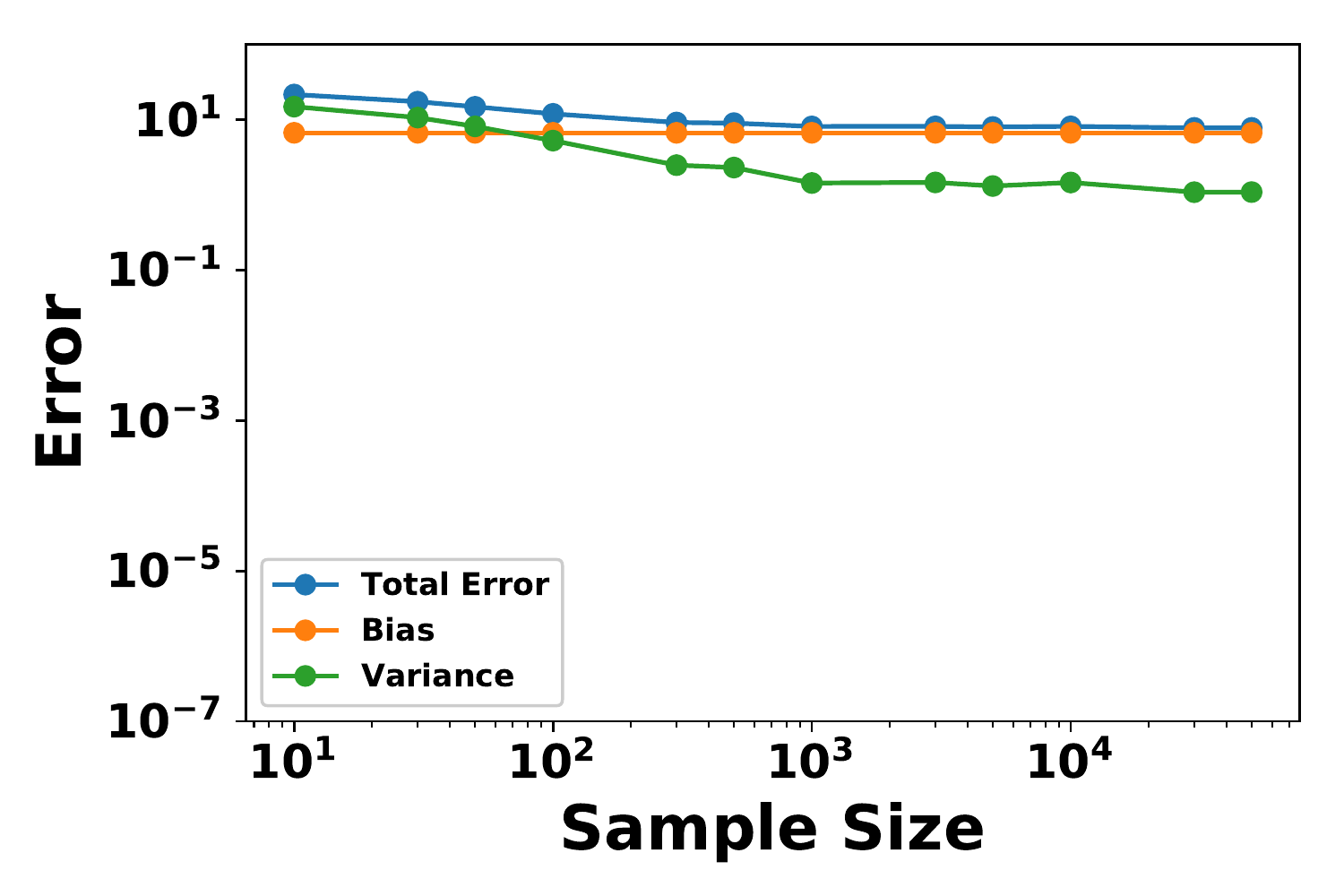}
              \caption{Order 7}
          \end{subfigure}
          \begin{subfigure}[b]{0.245\textwidth}
              \centering
              \includegraphics[width=\textwidth]{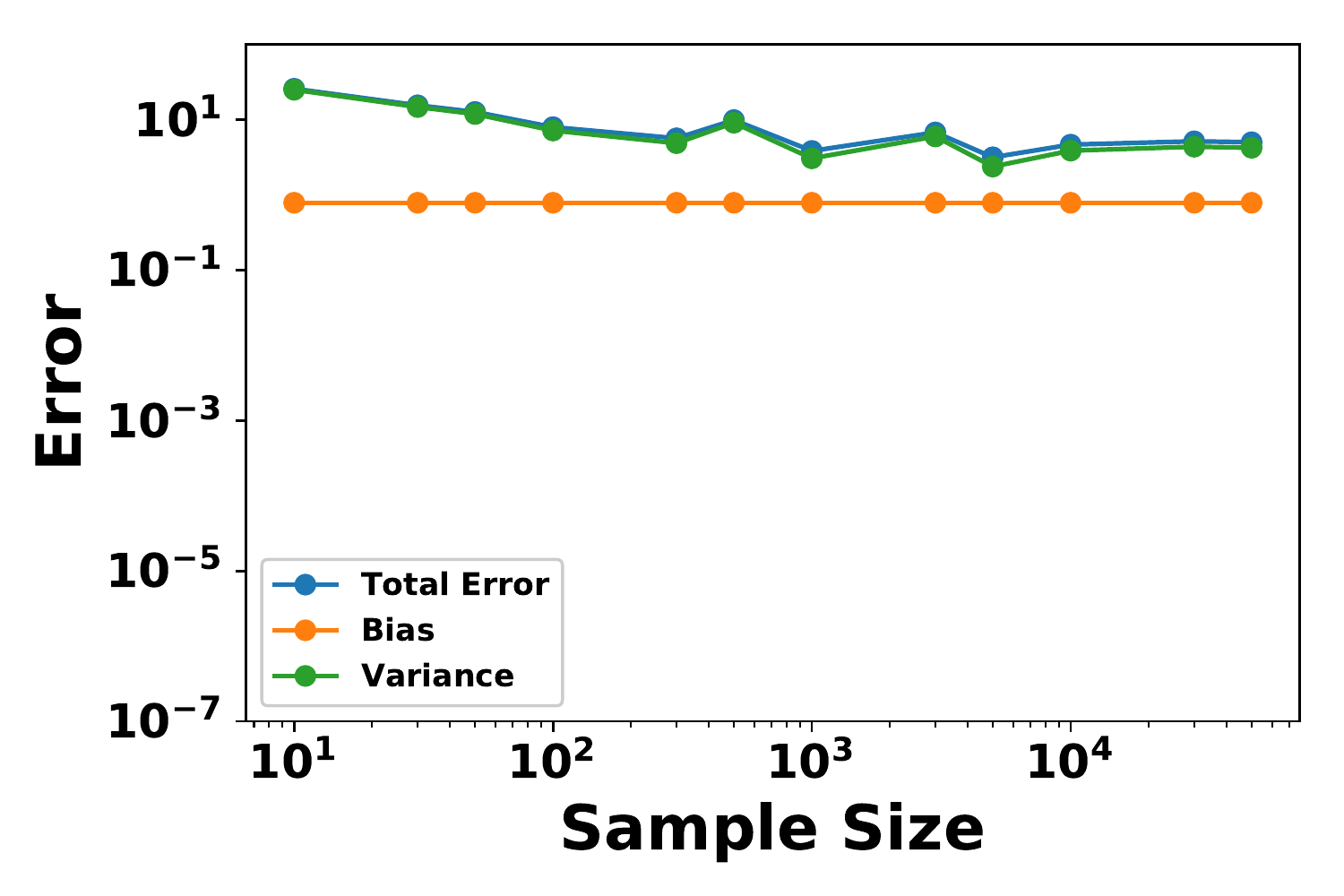}
              \caption{Order 10}
          \end{subfigure}
          \caption{Empirical evaluation of the error generated from the bias and variance for varying order of interactions in the HBM} \label{fig:HBM_sample_size}
        \end{figure*}

        \begin{figure*}[p]
          \centering
          \begin{subfigure}[b]{0.245\textwidth}
              \centering
              \includegraphics[width=\textwidth]{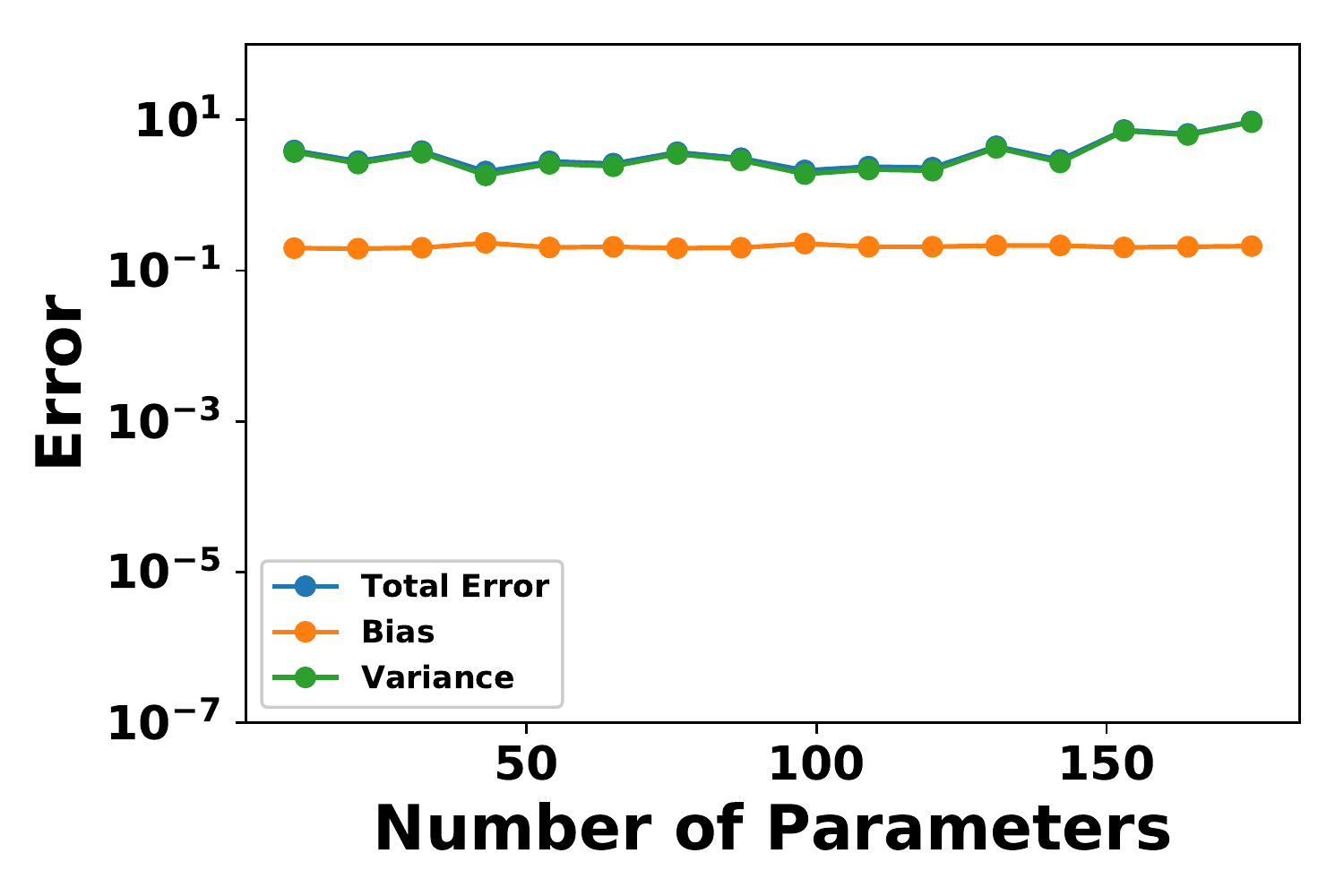}
              \caption{50 Samples}
          \end{subfigure}
          \begin{subfigure}[b]{0.245\textwidth}
              \centering
              \includegraphics[width=\textwidth]{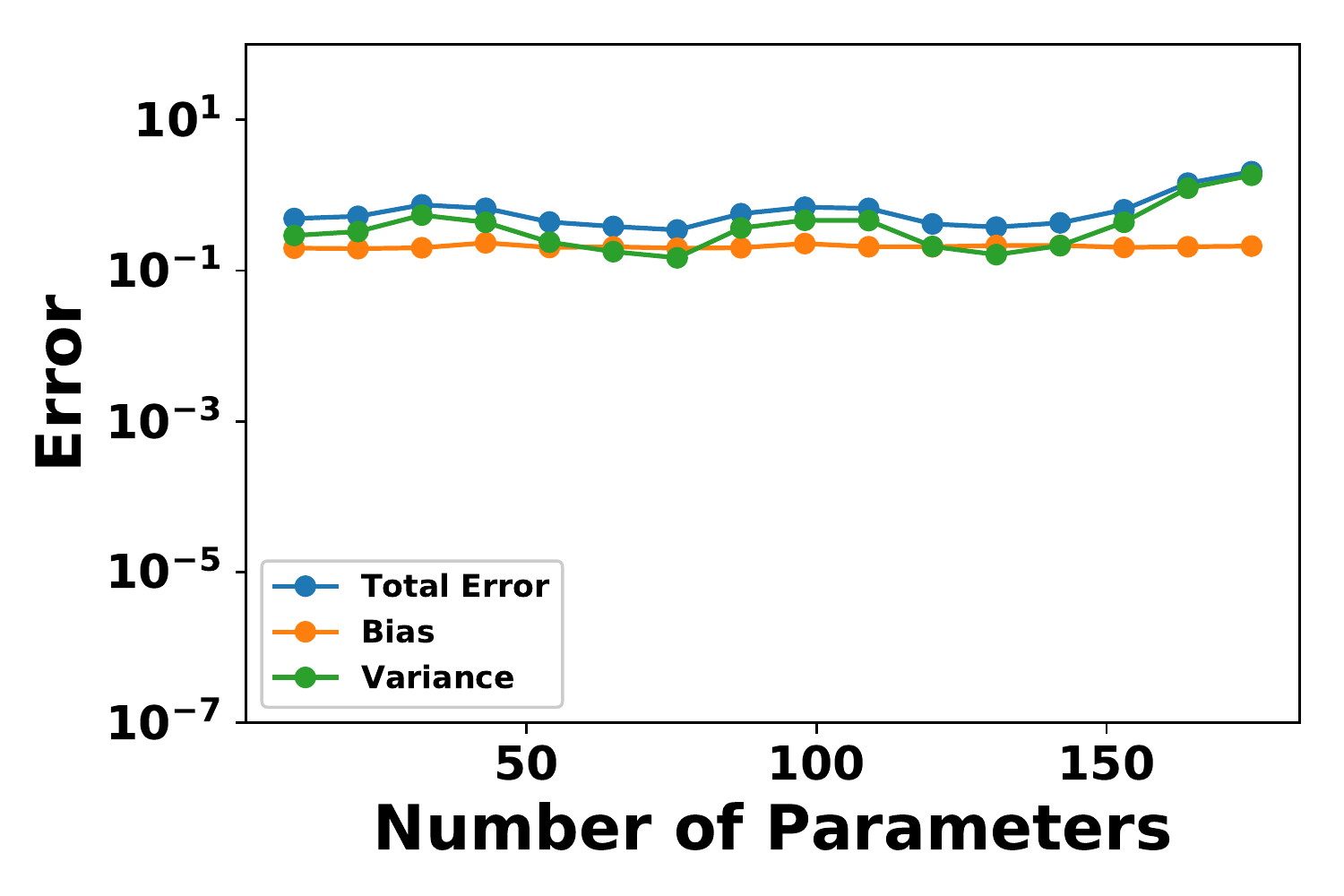}
              \caption{500 Samples}
          \end{subfigure}
          \begin{subfigure}[b]{0.245\textwidth}
              \centering
              \includegraphics[width=\textwidth]{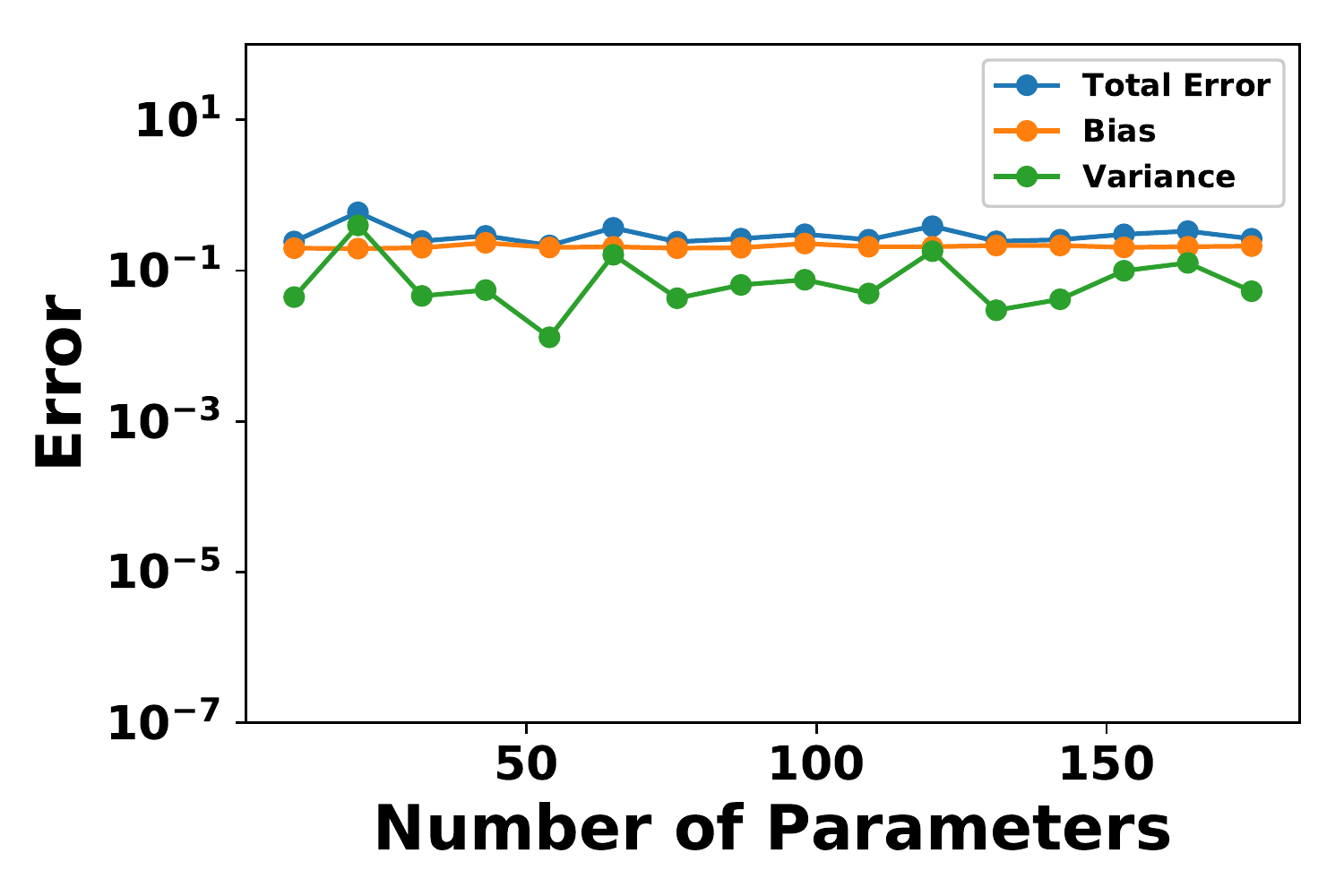}
              \caption{5000 Samples}
          \end{subfigure}
          \begin{subfigure}[b]{0.245\textwidth}
              \centering
              \includegraphics[width=\textwidth]{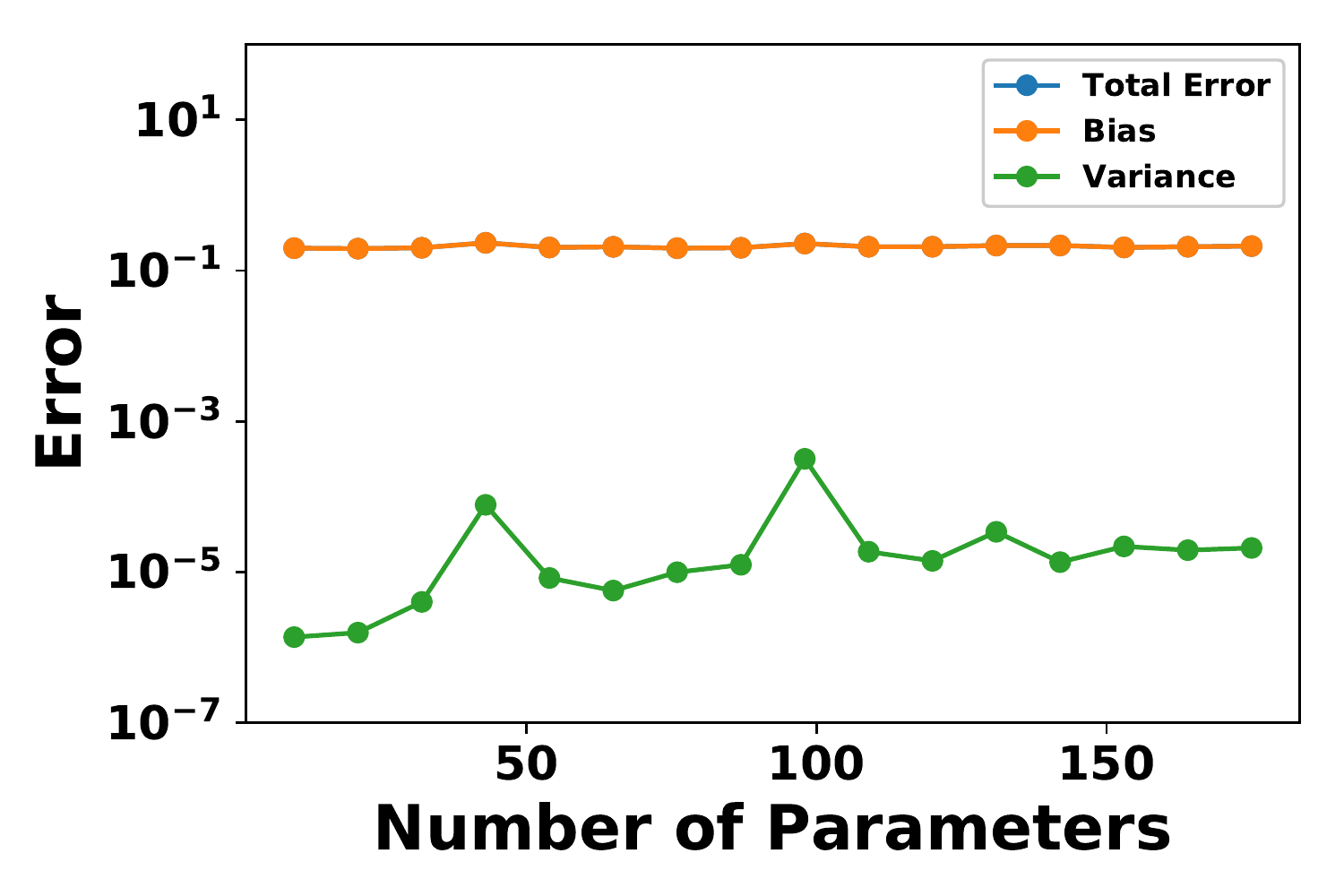}
              \caption{50000 Samples}
          \end{subfigure}
          \caption{Empirical evaluation of the error generated from the bias and variance for varying sample size in the RBM} \label{fig:RBM_hidden_nodes}
        \end{figure*}

        \begin{figure*}[p]
          \centering
          \begin{subfigure}[b]{0.245\textwidth}
              \centering
              \includegraphics[width=\textwidth]{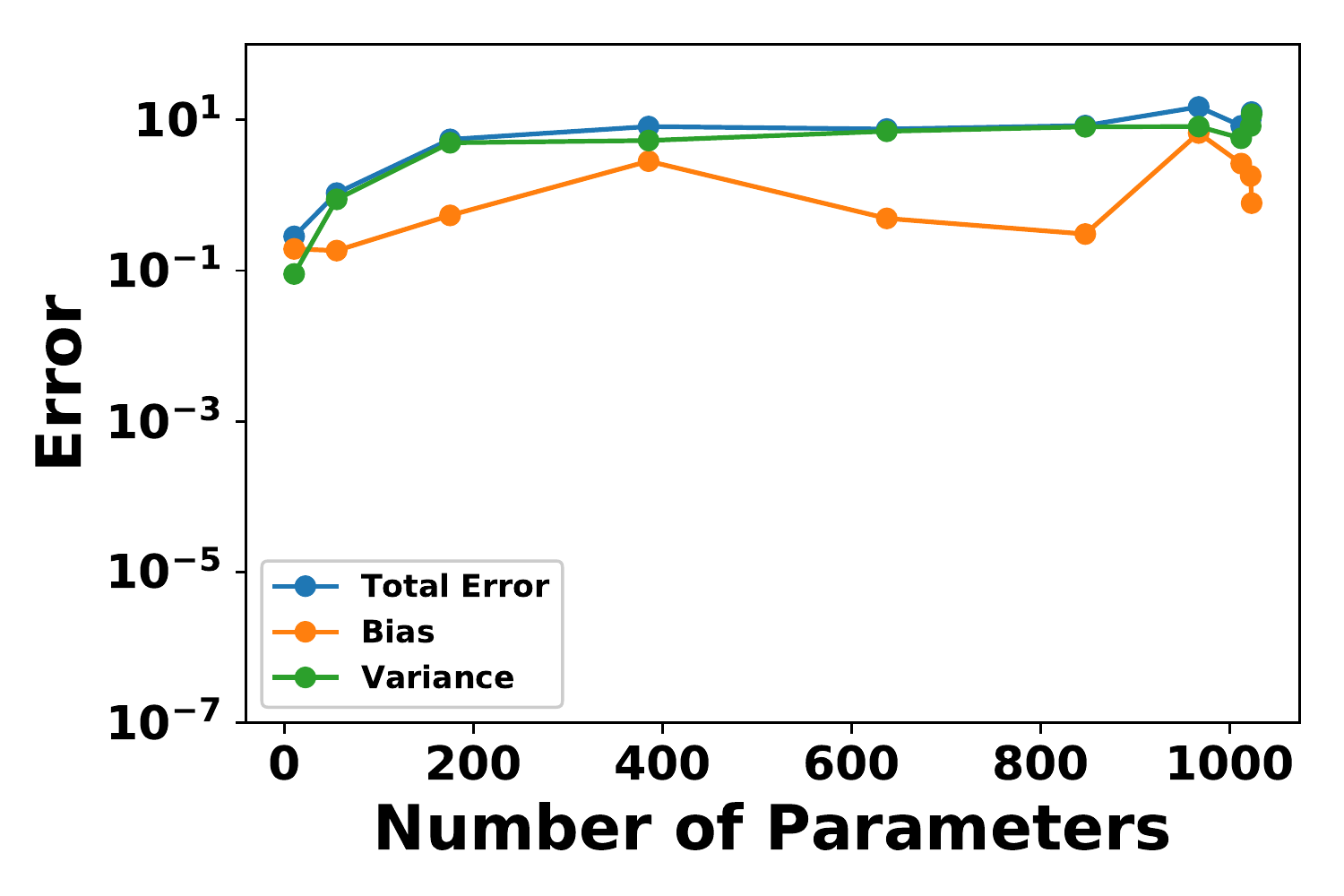}
              \caption{50 Samples}
          \end{subfigure}
          \begin{subfigure}[b]{0.245\textwidth}
              \centering
              \includegraphics[width=\textwidth]{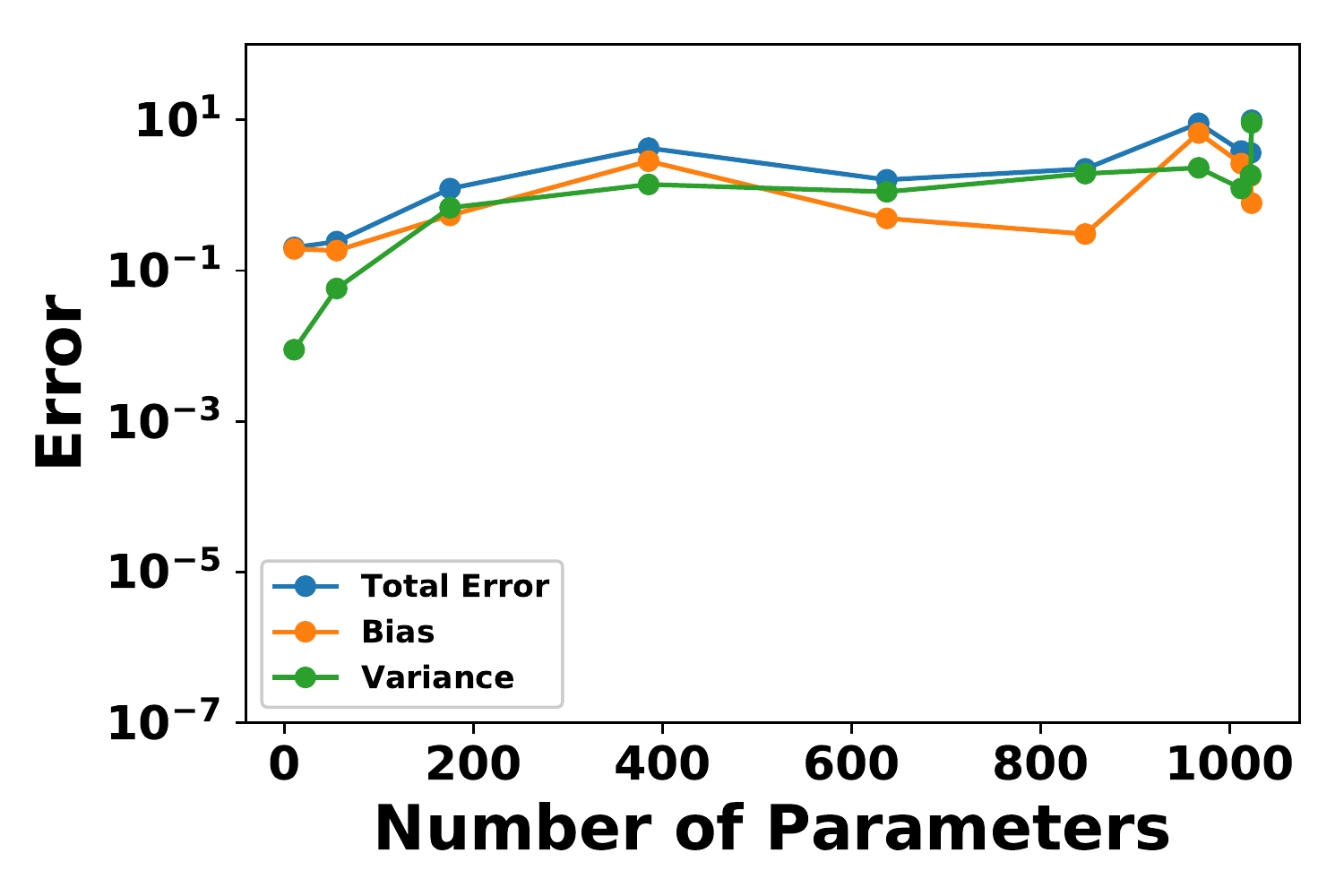}
              \caption{500 Samples}
          \end{subfigure}
          \begin{subfigure}[b]{0.245\textwidth}
              \centering
              \includegraphics[width=\textwidth]{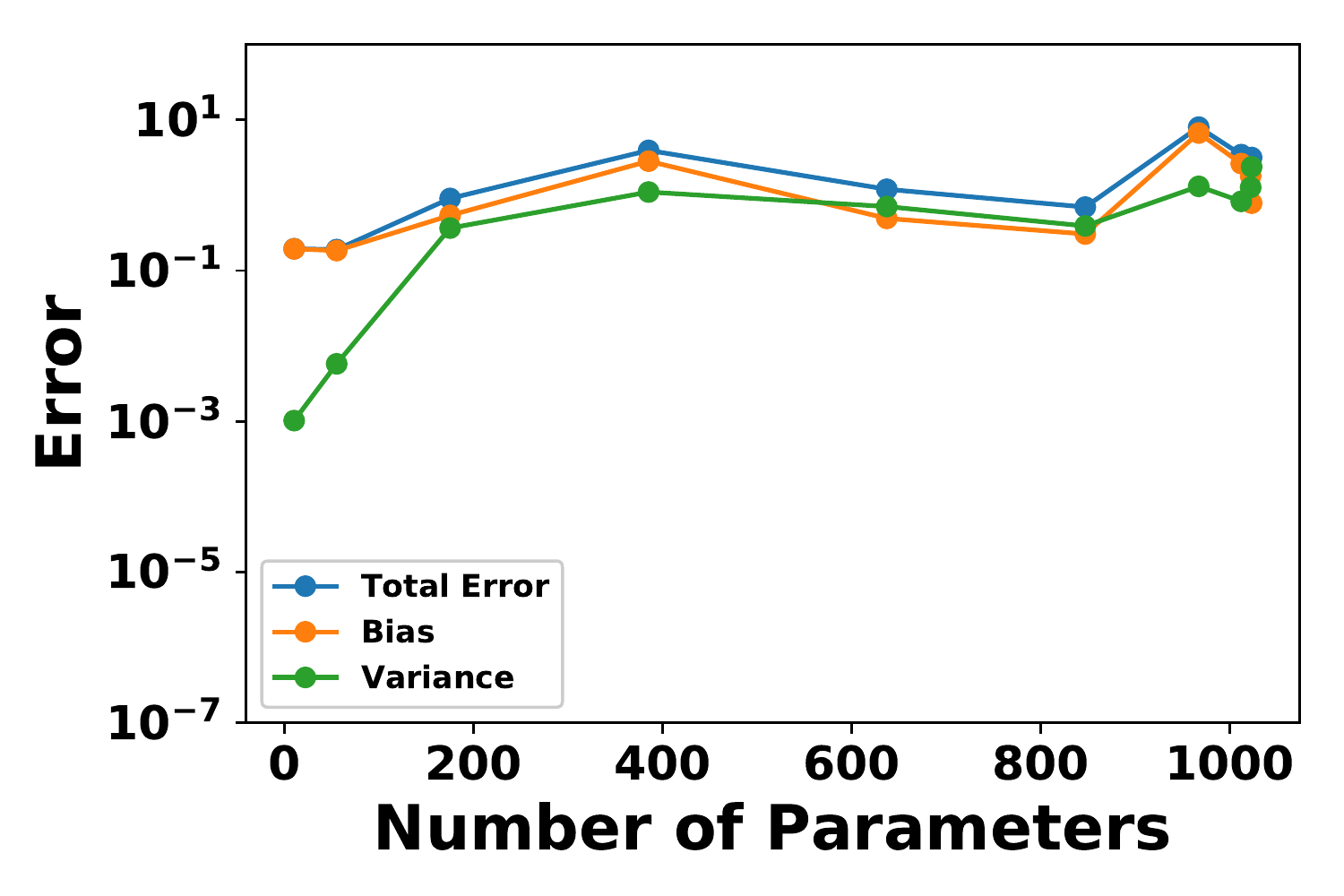}
              \caption{5000 Samples}
          \end{subfigure}
          \begin{subfigure}[b]{0.245\textwidth}
              \centering
              \includegraphics[width=\textwidth]{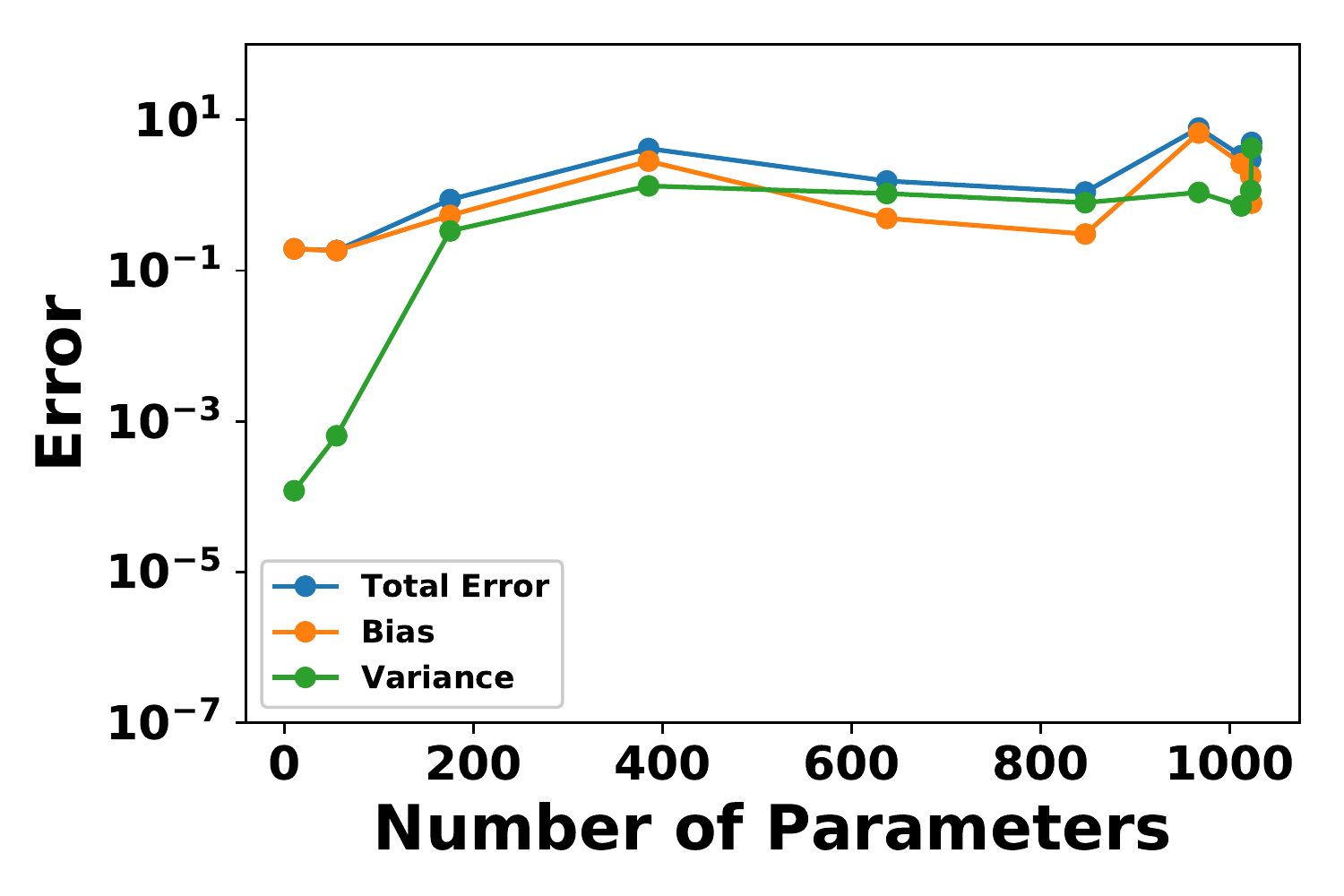}
              \caption{50000 Samples}
          \end{subfigure}
          \caption{Empirical evaluation of the error generated from the bias and variance for varying sample size in the HBM} \label{fig:HBM_hidden_nodes}
        \end{figure*}

	\subsection{Experiment Setup}
		A synthetic dataset is generated to study the bias-variance trade-off in HBM and RBM. The synthetic data is created by drawing a random probability $\left[ 0, 1 \right]$ from a uniform distribution for each $P^{*} \in \mathcal{P}^{*}$ such that $\sum_{P^{*} \in \mathcal{P}^{*}} P^{*} = 1$, where $\mathcal{P}^{*}$ represents the set of probabilities for all possible feature combinations. A sample size of $N =$ $\{$ $1 \times 10$, $3 \times 10$, $5 \times 10$, $1 \times 10^2$, $3 \times 10^2$, $5 \times 10^2$, $1 \times 10^3$, $3 \times 10^3$, $5 \times 10^3$, $1 \times 10^4$, $3 \times 10^4$, $5 \times 10^4$ $\}$ are drawn from the descrete probability distribution $\mathcal{P}^{*}$ using a multinomial. For each sample size $N$, we create 24 independent datasets to be used for both the HBM and RBM. The MLE of the HBM is calculated analytically by directly placing the true probability distribution $P^{*}$ into the model. While, for the RBM, it is not analytically tractable to calculate the MLE, it is instead approximated by placing a dataset several orders of magnitude larger than the experimental dataset, in our case, we have generated a dataset with the sample size of $1 \times 10^6$ and have assumed this to be the MLE. Both the HBM and the RBM has been run using $10,000$ Gibbs samples, a learning rate of $0.1$ and $10,000$ iterations.

        \begin{figure*}[ht]
          \centering
          \begin{subfigure}[b]{0.495\textwidth}
              \centering
              \includegraphics[width=\textwidth]{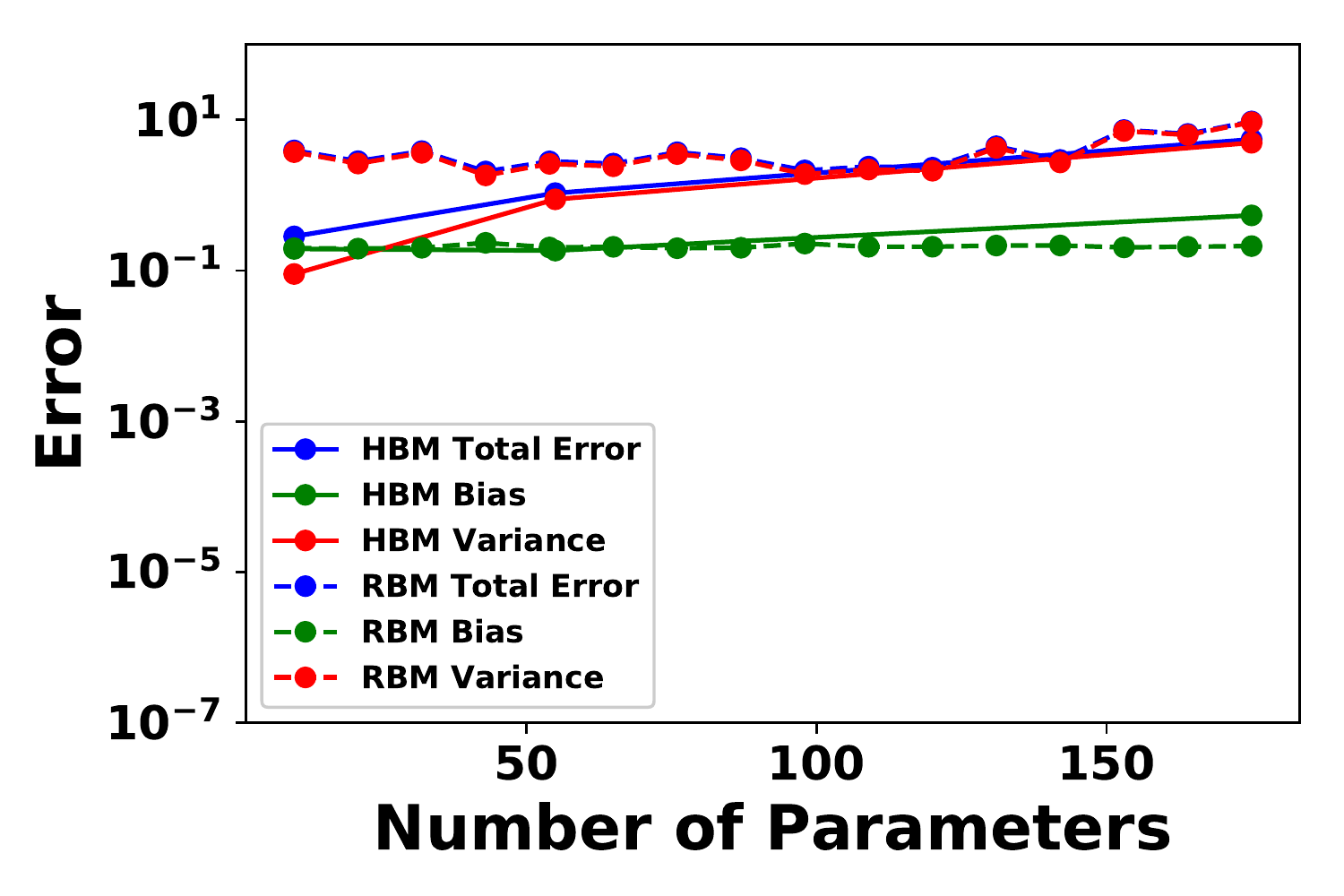}
              \caption{50 Samples}
          \end{subfigure}
          \begin{subfigure}[b]{0.495\textwidth}
              \centering
              \includegraphics[width=\textwidth]{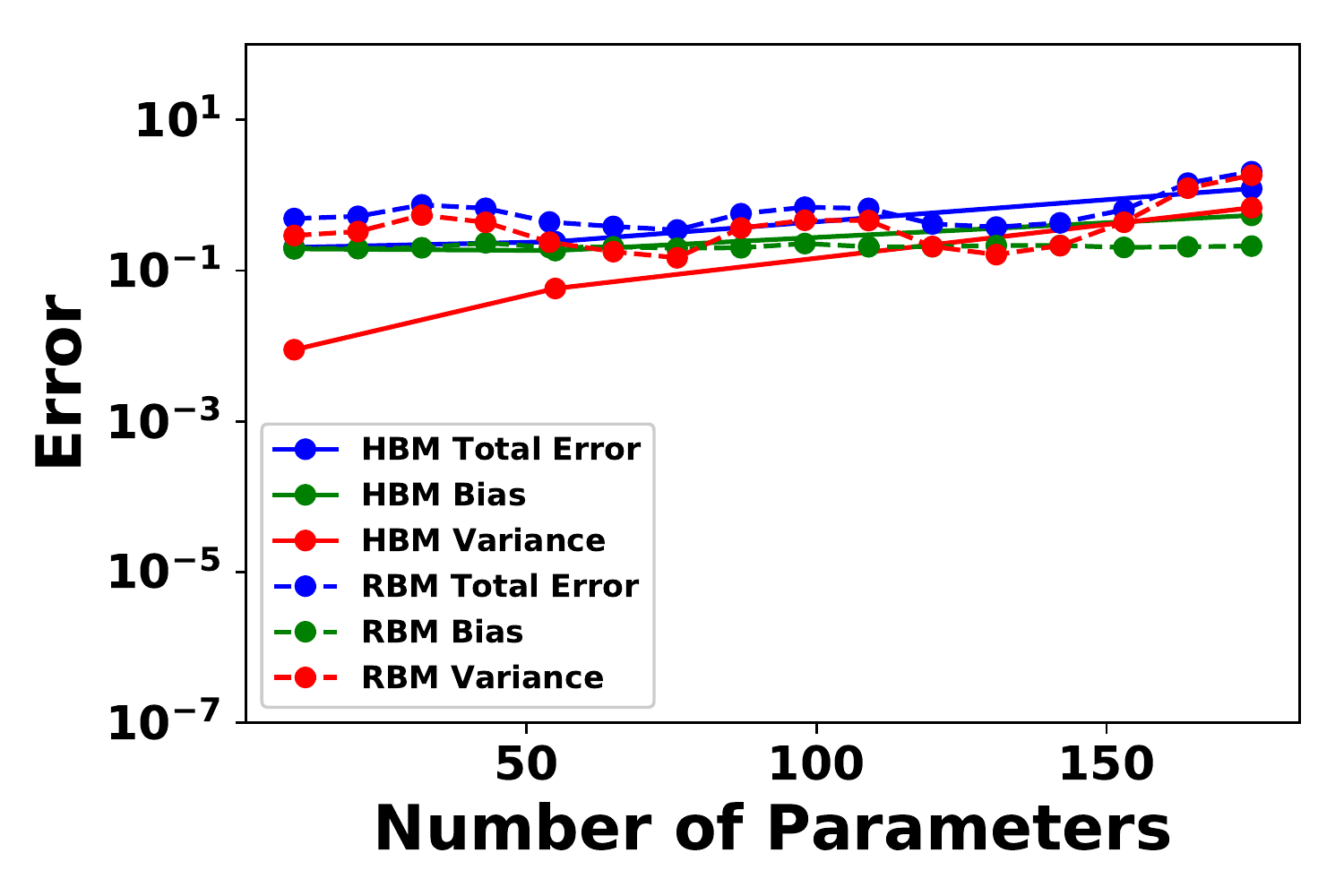}
              \caption{500 Samples}
          \end{subfigure}
          \begin{subfigure}[b]{0.495\textwidth}
              \centering
              \includegraphics[width=\textwidth]{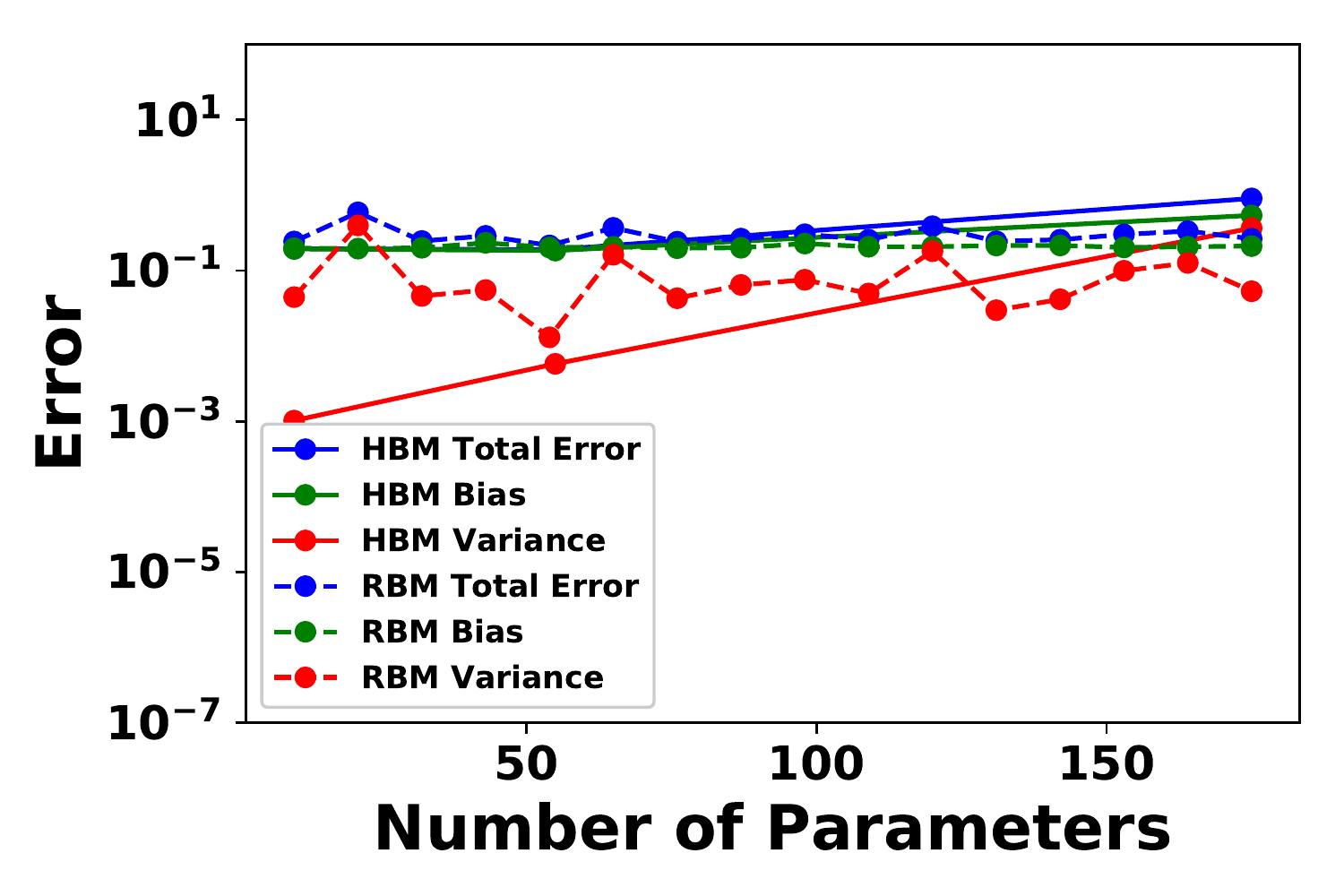}
              \caption{5000 Samples}
          \end{subfigure}
          \begin{subfigure}[b]{0.495\textwidth}
              \centering
              \includegraphics[width=\textwidth]{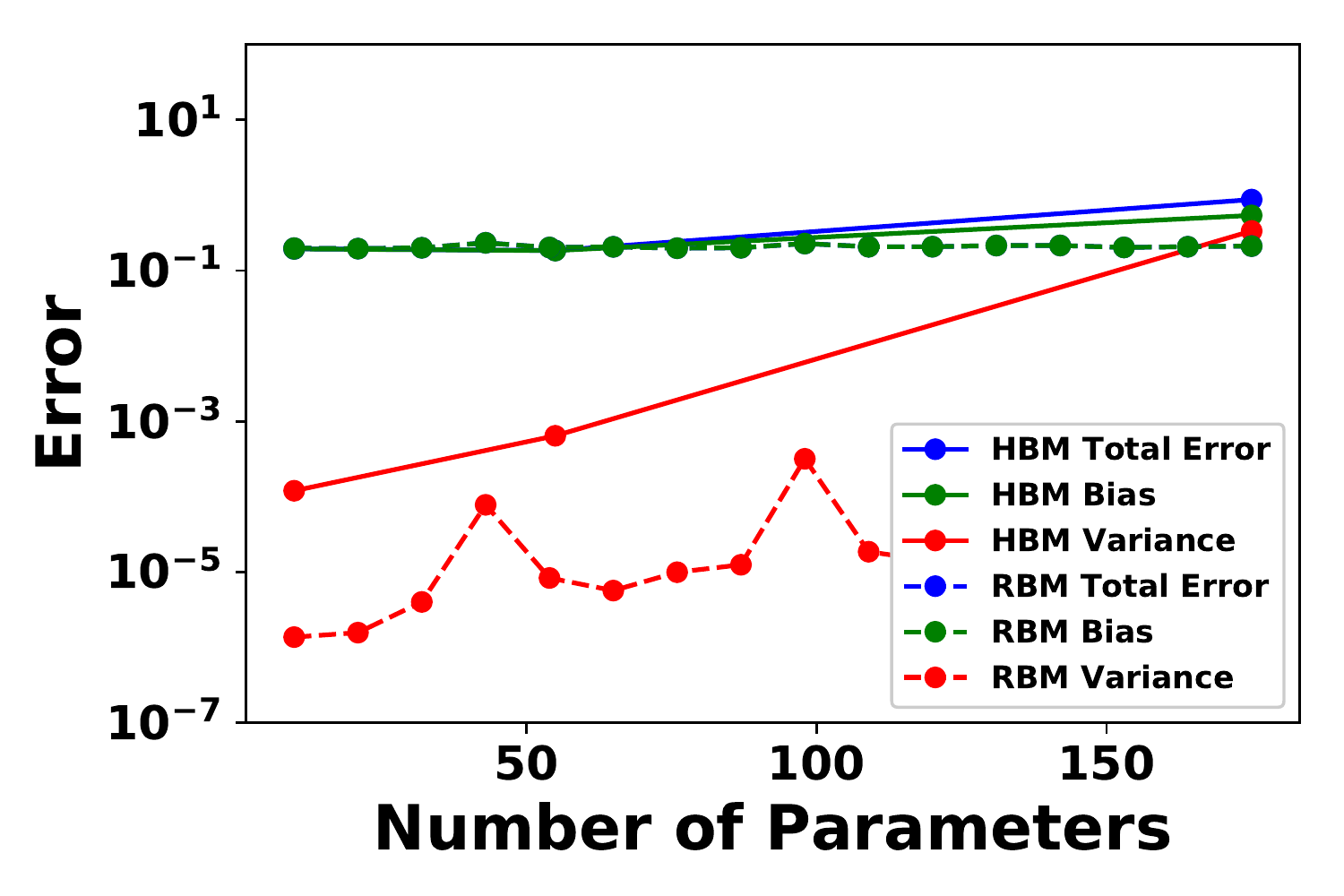}
              \caption{50000 Samples}
          \end{subfigure}
          \caption{Comparing empirical error in model for the HBM with RBM against the number of model parameters} \label{fig:RBM_vs_HBM}
        \end{figure*}

	\subsection{Experiment Results}
		The empirical evaluation in Figure~\ref{fig:RBM_error_plot} and Figure~\ref{fig:HBM_error_plot} have shown that both RBM and HBM have shown similar trends, with a positive correlation between the number of model parameters and variance and an inverse relationship between the sample size and variance. Surprisingly, the bias does not show any clear correlation between the number of model parameters and the sample size.
        
        The error generated from the variance is much more dominant for a smaller sample size and a larger number of model parameters. Comparing Figure~\ref{fig:RBM_sample_size} and Figure~\ref{fig:HBM_sample_size}, the RBM has shown to be more effective at reducing the variance with a larger sample size. The importance sampling used to estimate the partition function in HBM may have led to the higher variance in empirical results for the HBM.
        Figure~\ref{fig:RBM_hidden_nodes} and Figure~\ref{fig:HBM_hidden_nodes} shows a positive correlation between the model parameters and variance. The HBM shows to have a larger correlation between the model parameters and variance.
        
        The total error in the model is the sum of the bias and the variance. The total number of model parameters is a natural way to compare the total error generated by the RBM and HBM (i.e. $\left| \mathbf{b} \right| + \left| \mathbf{w} \right|$ for RBM  and $\left| \mathcal{S} \left( B \right) \right|$ for HBM). Figure~\ref{fig:RBM_vs_HBM} shows that for small sample size, HBM has shown to have produced a lower error in the model. The higher error in the RBM is generated by the larger variance. For a larger sample size, the error from the bias is more much dominate. Since the bias in both the RBM and HBM is in a similar order of magnitude, both models have a total error in the same order of magnitude. RBM has shown to be much more effective at reducing the variance with a larger sample size, however, this does not reduce the total error significantly because it is several orders of magnitude smaller than the bias.

\section{Conclusion} \label{sec:conclusion}
	In this paper, we have first proposed using a combination of Gibbs sampling and importance sampling to overcome the computational issues in training the information geometric formulation of the higher-order Boltzmann machine (HBM). The experimental results have shown that our proposed approach is effective in estimating the probability distribution of the model. Our proposed approach has been compared with the RBM to compare using \textit{hidden layers} and \textit{higher-order interactions} to model higher-order feature interactions. Our experimental results have shown that both models have produced a total error with similar orders of magnitude and using \textit{higher-order interactions} may be more effective at minimizing the variance for smaller sample size.

\bibliography{references}
\bibliographystyle{aaai}

\end{document}